\documentclass{article}

\PassOptionsToPackage{numbers, compress}{natbib}
\usepackage[preprint]{neurips_2025}

\usepackage[utf8]{inputenc} 
\usepackage[T1]{fontenc} 
\usepackage{hyperref} 
\usepackage{url} 
\usepackage{booktabs} 
\usepackage{amsfonts} 
\usepackage{nicefrac} 
\usepackage{microtype} 
\usepackage{xcolor} 

\usepackage{amsmath}
\usepackage[pdftex]{graphicx}
\usepackage{algorithm}
\usepackage{algpseudocode}
\usepackage{booktabs}
\usepackage{multirow}
\usepackage{tikz}
\usetikzlibrary{positioning,arrows.meta,calc,shapes.arrows}
\usepackage{amssymb}
\usepackage{subcaption}
\usepackage{adjustbox}
\usepackage{wrapfig}
\usepackage[most]{tcolorbox}

\title{Are Multimodal Large Language Models Ready for Omnidirectional Spatial
Reasoning?} 

\author{
$^\dagger$Zihao Dongfang$^{1}$ \quad $^\dagger$Xu Zheng$^{1,2}$ \quad Ziqiao Weng$^{5}$ \quad Yuanhuiyi Lyu$^{1}$ \quad \\ \textbf{Danda Pani Paudel$^{2}$} \quad \textbf{Luc Van Gool$^{2}$} \quad \textbf{Kailun Yang$^{4}$} \quad \textbf{Xuming Hu$^{1,3}$}\thanks{Corresponding author $^\dagger$Equal Contribution} \\
$^{1}$HKUST(GZ) \quad
$^{2}$INSAIT, Sofia University, St. Kliment Ohridski \quad \\
$^{3}$HKUST \quad
$^{4}$Hunan University \quad
$^{5}$Sichuan University \quad
}

\newcommand{\ie}{\textit{i.e.}}

\newcommand{\eg}{\textit{e.g.}}

\begin{document}
\maketitle
\vspace{-20pt}
\begin{figure}[h!]
\centering
\includegraphics[width=0.85\textwidth]{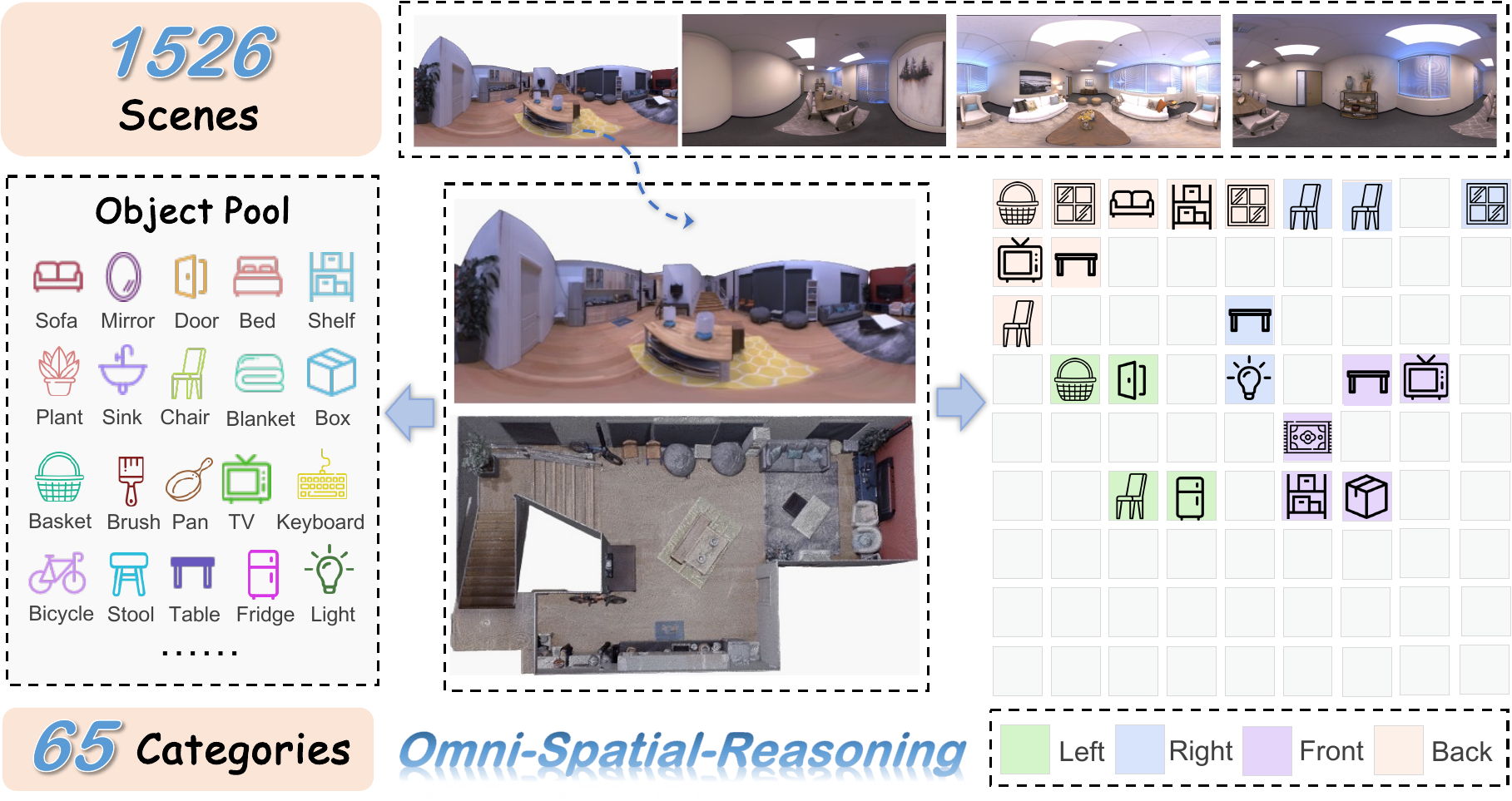}
\vspace{-4pt}
\caption{Overview of our \textit{\textbf{Omni-Spatial-Reasoning (OSR)}} dataset and benchmark. 
}
\label{fig:cover_figure}
\end{figure}


\begin{abstract}
The \textit{180{\textdegree}$\times$360{\textdegree}} omnidirectional field of view captured by 360{\textdegree} cameras facilitates their application across a broad spectrum of downstream tasks such as embodied AI and virtual reality. Despite recent progress in exploring the visual-spatial reasoning capabilities of Multimodal Large Language Models (MLLMs), most existing efforts focus on pinhole-view images, leaving omnidirectional perception underexplored. In this paper, we ask: \textit{Are MLLMs ready for omnidirectional spatial reasoning?} To address this, we introduce \textbf{\textit{OSR-Bench}}, the first benchmark specifically designed for omnidirectional spatial reasoning. OSR-Bench comprises over \textbf{$153K$} diverse Question-Answer (QA) pairs grounded in high-fidelity omni-cognitive maps extracted from panoramic indoor scenes, covering key reasoning categories including object counting, relative distance, and relative direction. We further propose a \textit{negative sampling strategy} that injects non-existent objects into prompts to systematically evaluate hallucination and grounding robustness. To support fine-grained analysis, we design a two-stage evaluation framework that assesses both cognitive map generation and QA accuracy using rotation-invariant matching and both rule-based and LLM-based metrics. We benchmark eight state-of-the-art MLLMs, including GPT-4o, Gemini 1.5 Pro, and open-source counterparts, under zero-shot settings. Our results reveal that current models exhibit limited spatial reasoning capabilities under panoramic settings, highlighting the need for more robust and perceptually grounded MLLMs. The established OSR-Bench and source code will be made publicly available at \url{https://huggingface.co/datasets/UUUserna/OSR-Bench}.
\end{abstract}

\section{Introduction}
\textit{Omnidirectional images}, captured using 360{\textdegree} cameras, offer a full 180{\textdegree}$\times$360{\textdegree} field of view, enabling comprehensive perception of surrounding environments. 
This panoramic imaging capability has driven the adoption of omnidirectional sensors in numerous downstream applications, including Augmented Reality (AR), Virtual Reality (VR), and scene understanding tasks~\cite{jiang2025dimer,liao2025benchmarking,zheng2025retrieval,lee2024all,lyu2025realrag,zheng2025distilling}. 
Unlike conventional pinhole cameras that provide a limited field of view, 360{\textdegree} cameras capture spatial context in their entirety, making them especially advantageous for embodied AI and spatial reasoning scenarios. Consequently, there has been increasing research interest in leveraging omnidirectional perception for fine-grained scene understanding, particularly in tasks such as semantic segmentation~\cite{zheng2024open,zhang2024behind,zhang2022bending,zheng2023both,zheng2023look,zheng2024semantics,zheng2024360sfuda++,zhang2024goodsam,zhang2024goodsam++,cai2024interact360,zhong2025omnisam}.

\begin{figure}
\centering
\includegraphics[width=\textwidth]{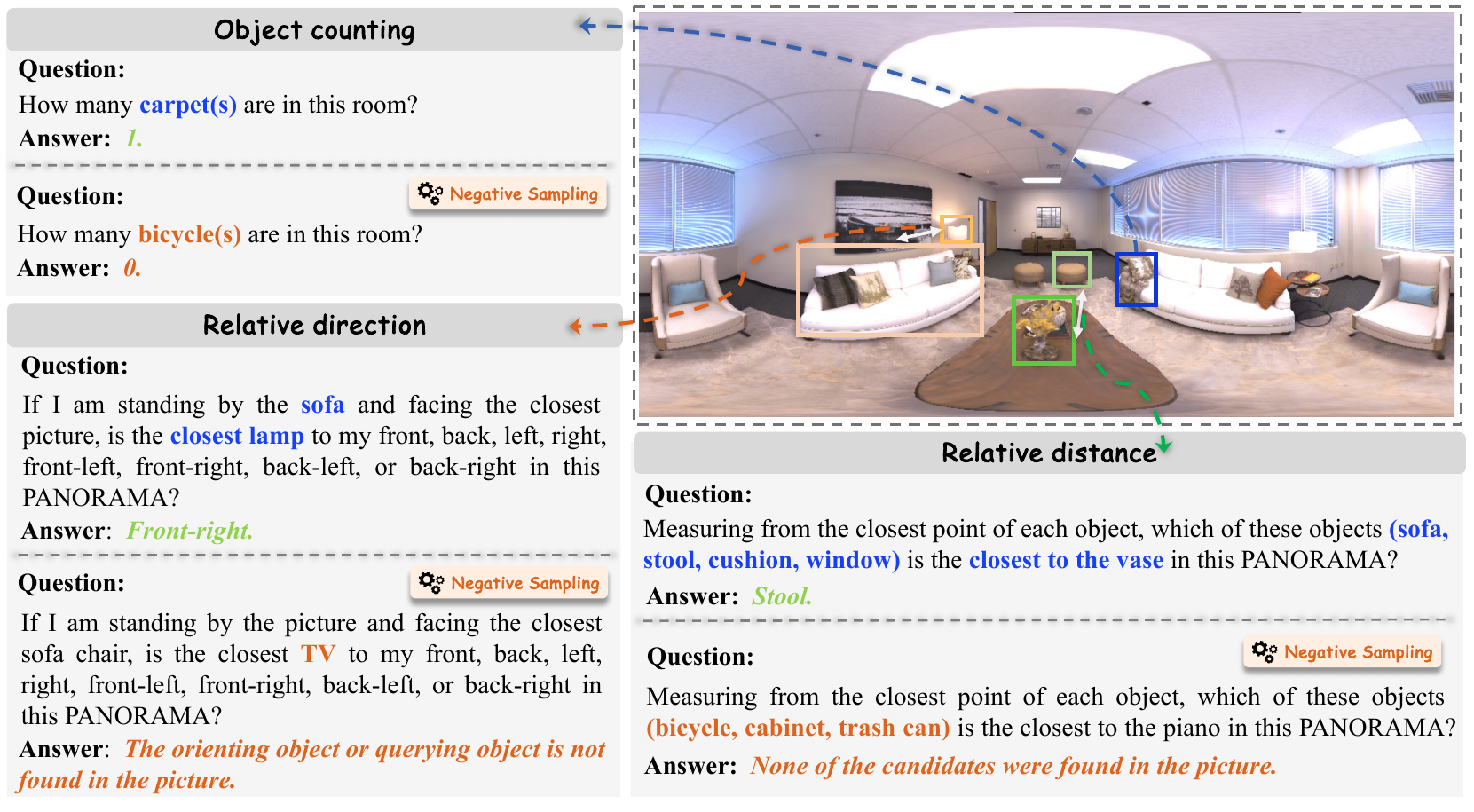}
\vspace{-16pt}
\caption{Cases of the established Omni-Spatial-Reasoning dataset.}
\vspace{-12pt}
\label{fig:QA_example}
\end{figure}
In parallel, the emergence of Large Language Models (LLMs) has brought transformative advances in multi-step reasoning, with growing efforts to extend these capabilities into the visual domain through Multimodal Large Language Models (MLLMs)~\cite{yan2024survey,huo2025mmunlearner,wang2024exploring,li2024manipllm}. While MLLMs have demonstrated remarkable performance across modalities such as vision, text, and audio, most existing approaches are built upon 2D planar images from pinhole cameras, thereby under-utilizing the rich spatial information encoded in omnidirectional views.
To bridge this gap, we present \textbf{OSR-Bench}—the first benchmark specifically designed to evaluate the spatial reasoning abilities of MLLMs in 360{\textdegree} panoramic environments. Built upon the ReplicaPano~\cite{dong2024panocontext} and DeepPanoContext~\cite{zhang2021deeppanocontext} datasets, OSR-Bench leverages high-fidelity 3D layout annotations to construct ground-truth top-down \textit{omni-cognitive maps}, which are used both as evaluation references and as structured priors for generating diverse spatial reasoning questions, as shown in Figure~\ref{fig:cover_figure}.

We develop a scalable, templated QA generation pipeline that produces question-answer pairs covering three key spatial reasoning categories: object counting, relative distance, and relative direction, with cases shown in Figure~\ref{fig:QA_example}. Each QA pair is derived from the panoramic image and its 3D layout, enabling direct grounding in visual geometry. To assess the robustness of visual understanding, we further introduce a \textit{negative sampling strategy} inspired by recent hallucination detection techniques~\cite{li2023evaluating}. By injecting non-existent distractor objects into prompts and QA templates, we generate both matched and mismatched question variants to systematically probe visual grounding fidelity. Our final benchmark comprises over \textbf{$153{,}000$} QA pairs across $4{,}100$ panoramic images from $1{,}526$ indoor scenes and $65$ object categories. OSR-Bench supports fine-grained, interpretable evaluation under both clean and adversarial conditions, offering a rigorous testbed for analyzing the spatial reasoning and robustness of MLLMs in realistic, panoramic settings.

To enable systematic and reproducible evaluation, we design a two-stage assessment protocol tailored for omnidirectional spatial reasoning. The first stage measures a model’s ability to construct spatially grounded cognitive maps from panoramic images, whereas the second evaluates its capacity to solve reasoning tasks such as object counting, relative distance, and direction. 
We adopt Hungarian matching with rotation-invariant F1 scores for cognitive map alignment, and employ both rule-based and LLM-based evaluators to assess question-answering performance, including hallucination sensitivity. We further evaluate a suite of eight state-of-the-art MLLMs spanning both proprietary (\eg, GPT-4o, Gemini 1.5 Pro) and open-source families (\eg, InternVL2.5~\cite{chen2025expandingperformanceboundariesopensource}, LLaVA-v1.5~\cite{liu2024improvedbaselinesvisualinstruction}, LLaMA3.2-Vision~\cite{grattafiori2024llama3herdmodels}, Qwen2.5-VL~\cite{bai2023qwenvlversatilevisionlanguagemodel}, DeepSeek-VL2~\cite{lu2024deepseekvlrealworldvisionlanguageunderstanding}, Janus-Pro~\cite{chen2025janusprounifiedmultimodalunderstanding}). All evaluations are conducted in strict zero-shot settings without any task-specific fine-tuning. To ensure methodological consistency and result reproducibility, we apply greedy decoding across all models.


In summary, our work makes the following contributions:
\textbf{(I)} We introduce \textbf{OSR-Bench}, the first benchmark specifically tailored for evaluating the spatial reasoning abilities of Multimodal Large Language Models (MLLMs) in 360{\textdegree} panoramic environments. \textbf{(II)} We develop a scalable data generation pipeline that constructs over \textbf{$153,000$} high-quality QA pairs across three spatial reasoning categories, grounded in omni-cognitive maps derived from photorealistic indoor layouts. \textbf{(III)} We propose a novel \textit{negative sampling strategy} that injects non-existent distractor objects into both prompts and QA templates, enabling robust evaluation of hallucination and visual grounding fidelity. \textbf{(IV)} We design a two-stage evaluation framework combining rotation-invariant cognitive map matching with both rule-based and LLM-based QA assessment, supporting interpretable and reproducible benchmarking. \textbf{(V)} We conduct a comprehensive evaluation across \textbf{8 state-of-the-art MLLMs}, both proprietary and open-source, under strict zero-shot settings, establishing the first comparative study of panoramic spatial reasoning in the multimodal era.

\section{Related Work and Motivation}  

\noindent\textbf{Visual Understanding and Reasoning.}  
Visual understanding and reasoning lie at the intersection of computer vision and natural language processing, with tasks like Visual Question Answering (VQA) serving as key benchmarks. Datasets such as VQAv2~\cite{goyal2017making}, GQA~\cite{hudson2019gqa}, and TextVQA~\cite{singh2019towards} have driven progress in this area by testing increasingly complex reasoning skills. More recent benchmarks like SEED-Bench~\cite{li2023seed}, MMBench~\cite{liu2024mmbench}, and MM-Vet~\cite{yu2024mm} extend these efforts to multimodal large language models (MLLMs) across diverse visual-linguistic tasks. For multi-turn dialogue, benchmarks like ConvBench~\cite{liu2024convbench} and MMDU~\cite{liu2024mmdu} further push the field by introducing hierarchical capabilities and multi-image contexts, respectively. However, challenges like object hallucination persist, as highlighted by POPE~\cite{li2023evaluating} and NOPE~\cite{lovenia2024negative}, which measure models' tendency to generate false object references. Spatial reasoning has also received attention, with datasets like What's-Up~\cite{wang2024picture} and VSI-Bench~\cite{yang2024thinking} evaluating how effectively models handle spatial relationships and cognitive mapping.  

\noindent\textbf{Motivation.}  
Despite these advancements, most existing benchmarks focus on standard field-of-view images, limiting their ability to assess spatial understanding in more complex environments. Notably, they overlook the unique challenges of omnidirectional perspectives, which involve handling object relationships, counting, and relative direction in panoramic scenes. To address this gap, we propose OSR-Bench, the first benchmark specifically designed to evaluate spatial reasoning in 360{\textdegree} environments, providing a more comprehensive assessment of MLLMs.

\noindent\textbf{Panoramic Scene Understanding.}  
Panoramic scene understanding presents unique challenges compared to conventional narrow-field vision. The equirectangular projection used for 360{\textdegree} images introduces geometric distortion, while the expanded field of view increases object density and spatial complexity, complicating scene interpretation. Early work~\cite{zhang2014panocontext,zhang2021deeppanocontext,dong2024panocontext} in this area focused on specific tasks like room layout reconstruction, recovering instance-level geometry from panoramic images. However, these approaches often produce task-specific outputs (\textit{e.g.}, corner coordinates, mesh vertices) that are not directly compatible with Multimodal Large Language Models (MLLMs).  
Recent efforts have expanded to fine-grained scene understanding, including semantic segmentation~\cite{zheng2024open,zhang2024behind,zhang2022bending,zheng2023both,zheng2023look,zheng2024semantics,zheng2024360sfuda++,zhang2024goodsam,zhang2024goodsam++,cai2024interact360,zhong2025omnisam}, aiming to capture richer semantic content. This growth has been accompanied by the creation of specialized panoramic datasets. For example, Pandora~\cite{xu2022pandora} focuses on orientation-aware object detection, while 360+x~\cite{chen2024360+} and R3DS~\cite{wu2024r3ds} provide multimodal annotations for comprehensive scene understanding. Large-scale outdoor datasets like KITTI-360~\cite{liao2022kitti}, DensePASS~\cite{ma2021densepass}, and the Waymo Open Dataset~\cite{mei2022waymo} target autonomous driving, while Matterport3D~\cite{chang2017matterport3d} and Stanford2D3D~\cite{armeni2017joint} offer richly annotated indoor scans for embodied intelligence.  

\noindent\textbf{Motivation.}  
Despite these advancements, existing datasets face several limitations when used to evaluate MLLMs under a unified VQA paradigm:  
\textcircled{1} \textit{Task mismatch:} Many datasets focus on geometric reconstruction or dense prediction, which are not naturally suited for VQA-style evaluation.  
\textcircled{2} \textit{Translation overhead:} Converting these tasks into natural language queries requires extensive post-processing, often losing geometric context.  
\textcircled{3} \textit{Limited spatial QA coverage:} Current omnidirectional VQA datasets (e.g., Pano-AVQA~\cite{yun2021pano}, VQA 360{\textdegree}~\cite{chou2020visual}) lack comprehensive spatial reasoning, particularly for object positions across the full 360{\textdegree} field of view.  
\textcircled{4} \textit{Handcrafted bias:} Manually constructed QA pairs may not capture the full complexity of panoramic spatial reasoning, increasing the risk of hallucination.  
To address these gaps, we propose \textbf{\textit{OSR-Bench}}, a benchmark specifically designed to systematically assess spatial reasoning in panoramic scenes, providing a foundational and open resource for the community.

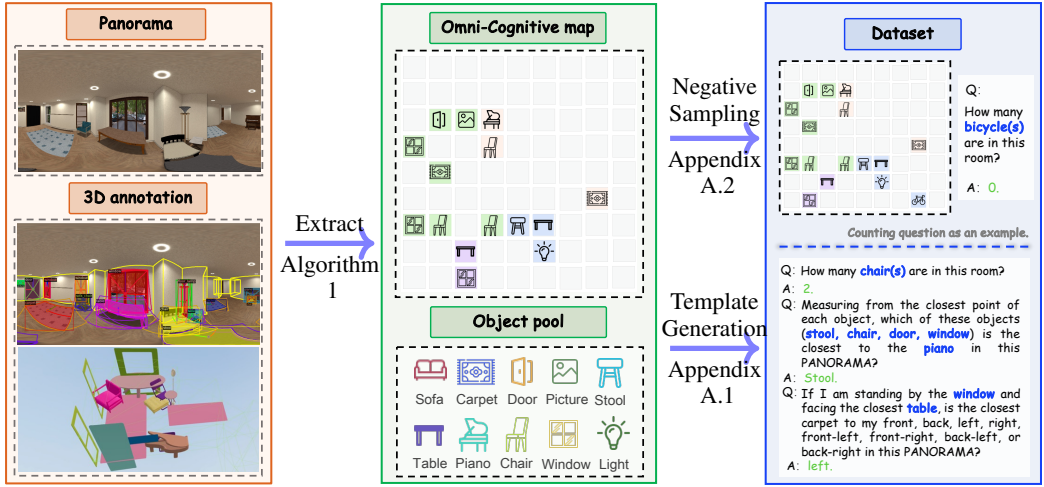
\begin{figure}
\centering
\begin{tikzpicture}[
    box/.style={rectangle, minimum width=0.26\textwidth, inner sep=0pt, align=center},
    blue-arrow/.style={
        ->,
        line width= 0.7mm,
        draw=blue!50,
        fill=blue!20,
        shorten >= 2pt,
        shorten <= 2pt
    },
    arrow-text/.style={font=\footnotesize, align=center},
    outer sep=0pt
]
\def\nodespacing{0.095\textwidth}
\def\offset{0.1\textwidth}
\node[box] (extract) {
    \includegraphics[width=0.256\textwidth]{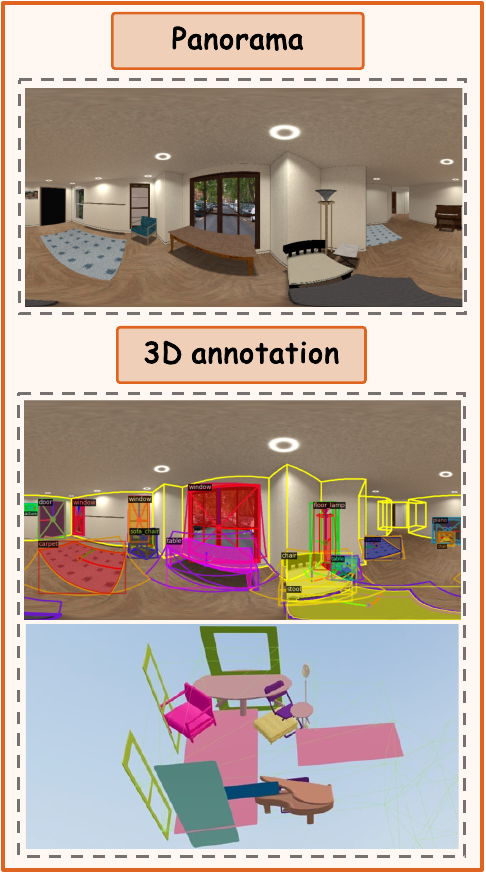}
};
\node[box, right=\nodespacing of extract] (sampling) {
    \includegraphics[width=0.265\textwidth]{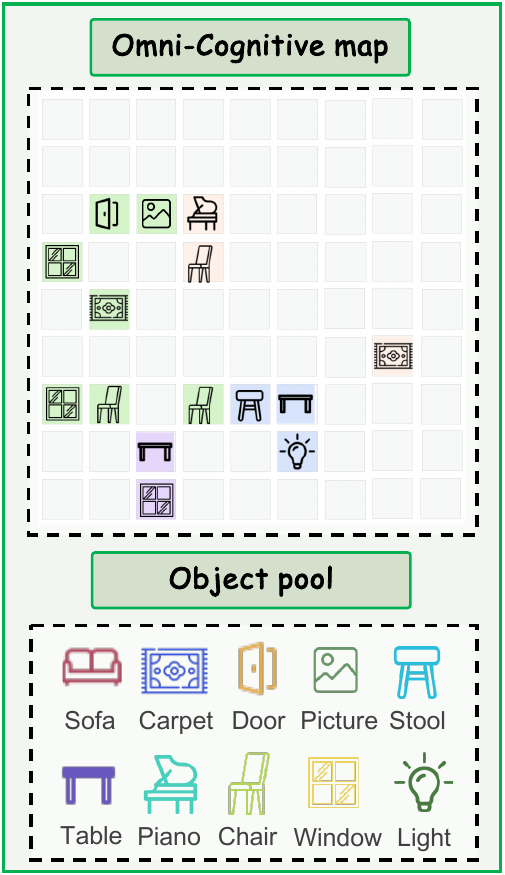}
};
\node[box, right=\nodespacing of sampling] (template) {
    \includegraphics[width=0.265\textwidth]{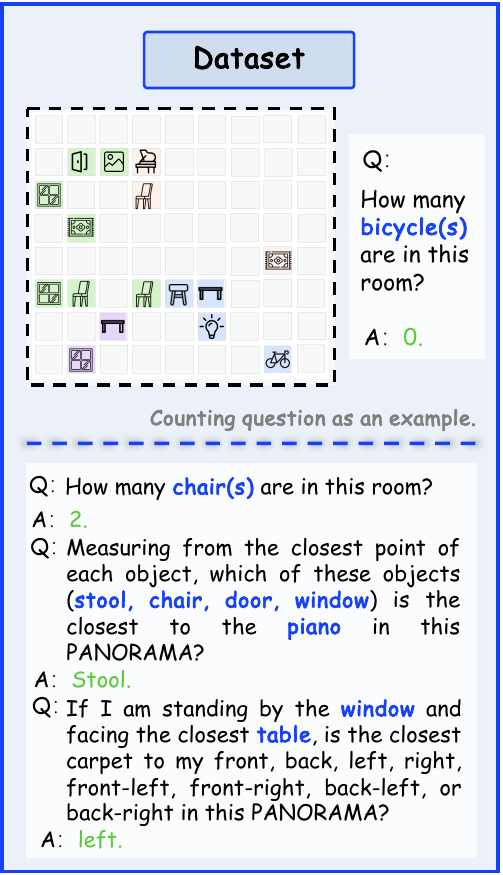}
};

\draw[blue-arrow] (extract.east) -- (sampling.west);
\node[arrow-text, above=0.05cm] at ($(extract.east)!0.5!(sampling.west)$) {Extract};
\node[arrow-text, below=0.05cm] at ($(extract.east)!0.5!(sampling.west)$) {Algorithm\\~\ref{alg:cogmap-extraction}};

\draw[blue-arrow] ($(sampling.east)+(0,\offset)$) -- ($(template.west)+(0,\offset)$);
\node[arrow-text, above=0.05cm] at ($(sampling.east)!0.5!(template.west)+(0,\offset)$) {Negative\\Sampling};
\node[arrow-text, below=0.05cm] at ($(sampling.east)!0.5!(template.west)+(0,\offset)$) {Appendix\\~\ref{alg:negative-sampling}};

\draw[blue-arrow] ($(sampling.east)-(0,\offset)$) -- ($(template.west)-(0,\offset)$);
\node[arrow-text, above=0.05cm] at ($(sampling.east)!0.5!(template.west)-(0,\offset)$) {Template\\Generation};
\node[arrow-text, below=0.05cm] at ($(sampling.east)!0.5!(template.west)-(0,\offset)$) {Appendix\\~\ref{alg:QAGeneration}};
\end{tikzpicture}
\vspace{-12pt}
\caption{The data generation pipeline of our dataset.}
\vspace{-12pt}
\label{fig:dataset_pipeline}
\end{figure}
\section{OSR-Dataset \& Benchmark}
To explore the spatial reasoning ability of existing MLLMs, we develop an omnidirectional question answering construction pipeline to effectively and automatically generate high-quality QA pairs at scale, as illustrated in Fig.~\ref{fig:dataset_pipeline}. The following is the detailed dataset construction pipeline.

\subsection{Dataset Construction} \label{sec:data_construction}
\textbf{Data Collection.} \label{para:data_collection}
Our dataset is primarily built upon two public panoramic scene understanding datasets: ReplicaPano-Dataset~\cite{dong2024panocontext} and the dataset introduced in DeepPanoContext~\cite{zhang2021deeppanocontext}. 
Both datasets are derived from iGibson synthetic environments~\cite{li2022igibson}, providing photorealistic indoor panoramic scenes with comprehensive layout annotations. These datasets are originally designed for room layout reconstruction tasks. They contain detailed ground truth information about object positions, orientations, and room boundaries, enabling robust omni-cognitive map generation that is critical to our benchmark.

\textbf{Omni-Cognitive Map Extraction.} \label{para:cogmap_extraction}
From the layout annotations in the source datasets, we extract ground truth omni-cognitive maps representing top-down views of each room, as shown in Figure~\ref{fig:dataset_pipeline}. These omni-cognitive maps serve as simplified representations of object locations within the environment, similar to the approach in~\cite{yang2024thinking}. These omni-cognitive maps serve two purposes: 
\textcircled{1} The omni-cognitive maps serve as \textbf{\textit{ground truth}} references for evaluating the spatial understanding capabilities of LLMs that have been prompted to construct cognitive maps prior to addressing spatial reasoning questions, and \textcircled{2} The omni-cognitive maps enable \textbf{\textit{automatic generation}} of diverse spatial reasoning questions with verifiable answers using predefined templates. More specifically, we begin with each panoramic image and its associated 3D layout annotations, from which we extract all object instances and construct a ground‐truth omni-cognitive map representing the top-down layout of that scene. The same extracted objects are simultaneously organized into an object pool. Both the omni-cognitive maps and the object pools are then fed into our QA template generation module, where predefined templates are applied to produce diverse spatial‐reasoning questions. 

\noindent \textbf{Negative Sampling.} 
Unlike the existing spatial reasoning benchmark VSI-bench~\cite{yang2024thinking}, which relies solely on ground-truth objects to construct QA templates and generate prompts for cognitive map, our approach addresses an important limitation: \textbf{\textit{models may prioritize knowledge from textual input while ignoring potential mismatches between visual and textual information}}, as well as \textbf{\textit{hallucination issues in visual information}}. Drawing inspiration from POPE~\cite{li2023evaluating}, we implement negative sampling strategies (Popular Sampling and Adversarial Sampling, see Appendix~\ref{alg:negative-sampling} for details) by introducing objects that do not exist in each scene. These interference items are incorporated into QA templates and prompt words that guide the model in cognitive map generation, resulting in two versions of our QA dataset. This methodology primarily evaluates a model's visual-spatial understanding capabilities within a single-image panoramic perspective, focusing on three problem categories: object counting, relative distance, and relative direction, as illustrated in Figure~\ref{fig:QA_example}.


\noindent \textbf{Dataset Overview.}
Our \textit{OSR-Bench} comprises $4,100$ panoramic images with a total of $65$ object categories sourced from $1,526$ photorealistic synthetic indoor scenes. For each image, we generate $5$ to $10$ regular QA pairs for each evaluation category described in the previous paragraph, alongside an equal number of negative sampling QA pairs. To ensure data quality, we implement rigorous controls over the sampling process for both regular and negative sampling QA pairs, thereby eliminating ambiguity and preventing duplication in QA pair generation. 
In total, our comprehensive benchmark contains approximately $153k$ QA pairs, which lays a robust foundation for evaluating panoramic spatial reasoning capabilities compared to other existing related benchmarks in Table~\ref{tab:benchmark-comparison}. See Appendix~\ref{app:data_detail} for more statistical insights.

\begin{figure}
\centering
\includegraphics[width=\textwidth]{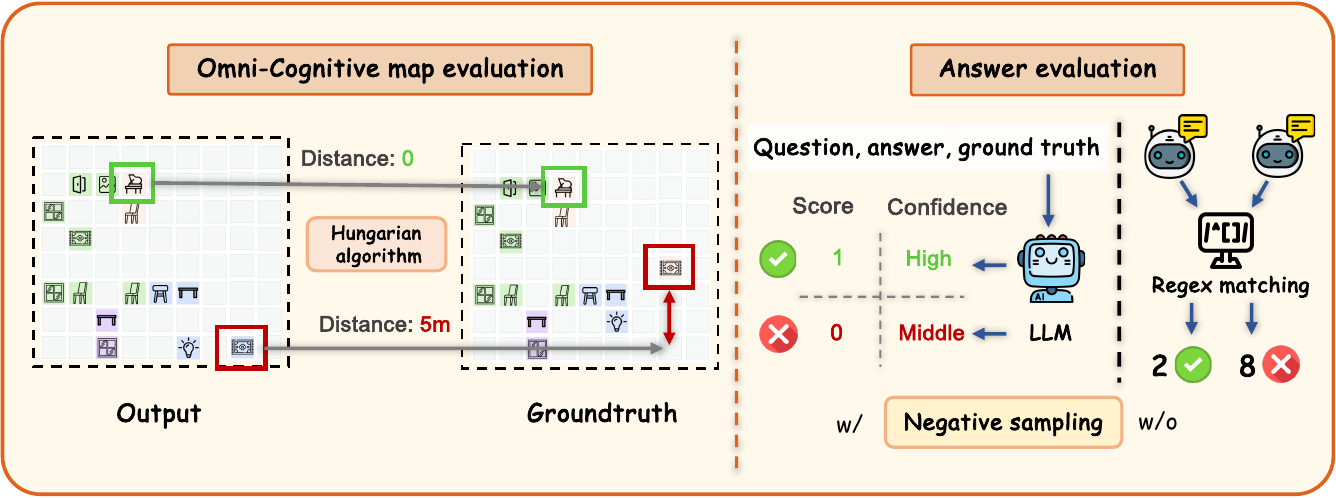}
\vspace{-16pt}
\caption{Benchmark Evaluation Pipeline.}
\label{fig:evaluation_pipeline}
\vspace{-12pt}
\end{figure}
\begin{table}[t!]
\centering
\caption{\textbf{Comparison of \textit{OSR-bench} with existing QA datasets. }
\textbf{Modality}: multimedia input data type; 
\textbf{Auto}: whether the production process is automatic or manual craft
\textbf{Dis}: presence of negative distractors by negative sampling to test hallucination;
\textbf{\#QAs}: number of QA pairs.}
\label{tab:benchmark-comparison}
\setlength\tabcolsep{1mm}
\resizebox{\linewidth}{!}{
\begin{tabular}{l l l c c r}
\toprule
Dataset & Modality & Domain & Auto & Dis & \#QAs\\
\midrule
Pano-AVQA~\cite{yun2021pano} & 360{\textdegree} videos~\& audio & simple spatial~\& audio reasoning & $\times$ & $\times$ & 51.7K \\
VQA 360{\textdegree}~\cite{chou2020visual} & 360{\textdegree} images & simple spatial~\& attributes reasoning & $\times$ & $\times$ & 16K \\
VIEW-QA~\cite{song2024videoquestion} & 360{\textdegree} videos~\& audio & daily tasks of the visually impaired & $\times$ & $\times$ & 4K \\
VSI-Bench~\cite{yang2024thinking} & NFOV videos & complex spatial reasoning & $\checkmark$ & $\times$ & 5K \\
\textit{OSR‐Bench(ours)} & \textbf{\textit{360{\textdegree} images}} & \textit{\textbf{complex spatial reasoning}} & $\checkmark$ & $\checkmark$ & \textbf{\textit{153K}} \\
\bottomrule
\end{tabular}}
\vspace{-12pt}
\end{table}

\subsection{Our Evaluation Method}
\label{evaluation_method}
OSR-Bench can be viewed as a two-round visual dialogue task. Accordingly, we adopt a two-stage evaluation framework tailored to omnidirectional spatial reasoning. The first stage measures a model's ability to construct accurate cognitive maps from panoramic images based on a comprehensive prompt. The second stage evaluates the model's capacity to answer specific spatial reasoning questions grounded in the constructed map. The overall evaluation pipeline is illustrated in Figure~\ref{fig:evaluation_pipeline}.

\noindent \textbf{Metrics.}
For cognitive map evaluation, we utilize the Hungarian algorithm to determine the optimal assignment between predicted and ground truth object locations. Given a predicted cognitive map $P$ and a ground truth cognitive map $G$, where $P_i$ represents the predicted positions of object class $i$ and $G_i$ represents the ground truth positions, we define a distance matrix $D^i \in \mathbb{R}^{|G_i| \times |P_i|}$ where each element $D^i_{jk}$ represents the Euclidean distance between the $j$-th ground truth position and the $k$-th predicted position for object class $i$:
\begin{equation}
\setlength{\abovedisplayskip}{3pt}
\setlength{\belowdisplayskip}{3pt}
D^i_{jk} = \sqrt{(G^i_j[0] - P^i_k[0])^2 + (G^i_j[1] - P^i_k[1])^2}.
\end{equation}
The Hungarian algorithm solves the assignment problem by finding the assignment matrix $X^i \in \{0,1\}^{|G_i| \times |P_i|}$ that minimizes the total distance:
\begin{equation}
\setlength{\abovedisplayskip}{3pt}
\setlength{\belowdisplayskip}{3pt}
\min_{X^i} \sum_{j=1}^{|G_i|} \sum_{k=1}^{|P_i|} D^i_{jk} X^i_{jk},
\end{equation}
subject to the constraints:
\begin{align}
\setlength{\abovedisplayskip}{3pt}
\setlength{\belowdisplayskip}{3pt}
\sum_{j=1}^{|G_i|} X^i_{jk} \leq 1, \quad \forall k \in \{1,2,\ldots,|P_i|\}, 
\sum_{k=1}^{|P_i|} X^i_{jk} \leq 1, \quad \forall j \in \{1,2,\ldots,|G_i|\}.
\end{align}
To accommodate potential rotational ambiguities in panoramic perception, we evaluate the cognitive map under rotational transformations and select the one that yields the highest F1 score. For each rotation $r$, we obtain a transformed prediction $P^r$ and compute:
\begin{equation}
\setlength{\abovedisplayskip}{3pt}
\setlength{\belowdisplayskip}{3pt}
F1^r = \frac{2 \times \text{Precision}^r \times \text{Recall}^r}{\text{Precision}^r + \text{Recall}^r},
\end{equation}
where Precision$^{r}$ and Recall$^{r}$ are calculated based on matches between $P^{r}$ and $G$ using a distance threshold of $2.0$ grid units.
For hallucination assessment, we use CHAIR (Confident Hallucination Assessment in Image Reasoning) metrics such as POPE~\cite{li2023evaluating}: CHAIR\_S, which indicates whether hallucination occurs in a scene(binary); CHAIR\_I, which measures the proportion of hallucinated classes among predicted classes; and CHAIR\_instance, which measures the proportion of hallucinated instances among all predicted instances. For regular spatial reasoning problems, we employ regex pattern matching to evaluate model responses against ground truth answers, which typically consist of concise words or phrases. Besides, we adopt category-specific evaluation metrics as follows:
\textcircled{\textbf{1}} For \textbf{\textit{Object counting}} tasks, we utilize a percentage-based accuracy measurement—awarding $100\%$ when the model's count exactly matches the ground truth, with scores decreasing proportionally as the numerical difference increases. 
\textcircled{\textbf{2}} For \textbf{\textit{Relative distance}} and \textbf{\textit{Relative direction}} tasks, we implement binary scoring (correct/incorrect) to assess response accuracy, given the discrete nature of these spatial relationships.

\noindent \textbf{LLM Evaluator}
For more complex evaluation scenarios, particularly with negative sampling, we adopt an LLM-based evaluator--which has been widely used in the evaluation of open-ended questions~\cite{zheng2023judging}, due to the challenge of evaluating answers about non-existent objects. The LLM evaluator provides binary scores ($0$ or $1$) and confidence levels (high, medium, low), providing detailed evaluation for ambiguous cases. Compared with traditional rule-based evaluation methods, it demonstrates robustness to linguistic variations in model outputs and maintains evaluation fairness across different models; compared with manual assessment methods, this method is more cost-effective and achieves greater objectivity and reduces bias. The specific prompts used for the LLM evaluator to evaluate each spatial reasoning task are detailed in the Appendix~\ref{prop:llm_evalautor}.

\begin{algorithm}[t!]
\caption{Omni-Cognitive Map Extraction}
\label{alg:cogmap-extraction}
\begin{algorithmic}[1]
\Require Scene annotations $S$, Grid size $G = 10 \times 10$
\State Extract layout boundaries $(x_{min}, y_{min}, x_{max}, y_{max})$ from $S$
\State Initialize cognitive map $C$ as $G \times G$ grid
\State Define mapping function: $(x, y) \mapsto (i, j)$ where $i \leftarrow \lfloor \frac{x - x_{min}}{x_{max} - x_{min}} \cdot G \rfloor$, $j \leftarrow \lfloor \frac{y - y_{min}}{y_{max} - y_{min}} \cdot G \rfloor$
\For{each object $o$ with centroid $(x, y)$ in $S$}
\State $(i, j) \leftarrow$ mapping of $(x, y)$ to grid coordinates using the defined function
\State $C[i][j] \leftarrow C[i][j] \cup {o.\text{class}}$  //Add object class to grid cell
\EndFor
\State $\mathcal{D} \leftarrow$ dictionary where $\mathcal{D}[c]$ is the count of class $c$ in $C$
\State \Return $C$, $\mathcal{D}$
\end{algorithmic}
\end{algorithm}

\subsection{Benchmark Experiments}
We comprehensively evaluate eight MLLMs across diverse model families. For proprietary models, we consider Gemini-1.5-Pro~\cite{geminiteam2024gemini15unlockingmultimodal} and GPT-4o~\cite{openai2023gpt4}. For open-source models, we evaluate models from InternVL2\_5-78B~\cite{chen2025expandingperformanceboundariesopensource}, LLaVA-v1.5-13B~\cite{liu2024improvedbaselinesvisualinstruction}, Llama-3.2-90B-Vision-Instruct~\cite{grattafiori2024llama3herdmodels}, Qwen2.5-VL-72B-Instruct~\cite{bai2023qwenvlversatilevisionlanguagemodel}, deepseek-vl2~\cite{lu2024deepseekvlrealworldvisionlanguageunderstanding} and Janus-Pro-7B~\cite{chen2025janusprounifiedmultimodalunderstanding}. To ensure methodological rigor and comparative validity, all evaluations are conducted under zero-shot settings without any task-specific fine-tuning. Furthermore, we employ greedy decoding across all models to maximize the reproducibility of our experimental results. See Appendix~\ref{evaluation_details} for details.

\subsubsection{Results}
\noindent \textbf{Cognitive Map Generation Capabilities.}

\begin{wraptable}{r}{10cm}
\centering
\vspace{-12pt}
\caption{\textbf{Omni-Cognitive Map Generation Performance.} Model performance on generating omni-cognitive maps from benchmark images with identical prompts, without negative sampling. \textbf{Best} results are in bold, and \underline{second-best} are underlined.}
\label{tab:omni-cogmap_Generation_comparison}
\renewcommand{\tabcolsep}{4pt}
\resizebox{\linewidth}{!}{
\begin{tabular}{lcccc}
\toprule
\textbf{Model} & \textbf{Avg distance} $\downarrow$ & \textbf{Precision} & \textbf{Recall} & \textbf{F1 score} \\
\midrule
GPT-4o~\cite{openai2023gpt4} & $\underline{3.501}$ & $\textbf{0.333}$ & $\underline{0.219}$ & $\underline{0.259}$ \\
Gemini-1.5-Pro~\cite{geminiteam2024gemini15unlockingmultimodal} & $\textbf{3.475}$ & $\underline{0.313}$ & $\textbf{0.238}$ & $\textbf{0.267}$ \\
Llama-3.2-90B-Vision-Instruct~\cite{grattafiori2024llama3herdmodels} & $5.991$ & $0.256$ & $0.188$ & $0.212$ \\
Qwen2.5-VL-72B-Instruct~\cite{bai2023qwenvlversatilevisionlanguagemodel} & $4.071$ & $0.307$ & $0.199$ & $0.231$ \\
InternVL2\_5-78B~\cite{chen2025expandingperformanceboundariesopensource} & $5.436$ & $0.291$ & $0.218$ & $0.243$ \\
LLaVA-v1.5-13B~\cite{liu2024improvedbaselinesvisualinstruction} & $8.512$ & $0.184$ & $0.153$ & $0.167$ \\
deepseek-v12~\cite{lu2024deepseekvlrealworldvisionlanguageunderstanding} &$9.607$ & $0.133$ & $0.146$ & $0.124$ \\
Janus-Pro-7B~\cite{chen2025janusprounifiedmultimodalunderstanding} & $14.467$ & $0.111$ & $0.131$ & $0.120$ \\
\bottomrule
\end{tabular}}
\vspace{-8pt}
\end{wraptable}
As shown in Table~\ref{tab:omni-cogmap_Generation_comparison}, proprietary models generally outperform open-source alternatives in generating omni-cognitive maps from panoramic images. Gemini-1.5-Pro achieves the highest F1 score of $0.267$, followed by GPT-4o at $0.259$. Among open-source models, InternVL2-5-78B performs best with an F1 score of $0.243$, approaching the proprietary models' results. However, there is a substantial gap between model families, with the lowest-performing model (Janus-Pro-7B) scoring only $0.008$. Notably, all models exhibit poor performance, with the best precision and recall values reaching only $0.333$ and $0.238$, respectively, highlighting the challenges in accurately capturing spatial arrangements in omnidirectional contexts.

\begin{table}[t!]
\caption{\textbf{Comparison of the average scores of the models for each type of question without negative sampling.} ``w/ cogmap'' indicates models were first prompted to generate a cognitive map of the omnidirectional scene prior to answering spatial questions, while ``w/o cogmap'' indicates models answered questions directly without the cognitive mapping step.}
\label{tab:comparison_withcogmap}
\centering
\resizebox{\textwidth}{!}{%
\begin{tabular}{l cc cc cc}
\toprule
\multirow{2}{*}{\textbf{Model}} & \multicolumn{2}{c}{\textbf{object count}} & \multicolumn{2}{c}{\textbf{relative distance}} & \multicolumn{2}{c}{\textbf{relative direction}} \\
\cmidrule(lr){2-3} \cmidrule(lr){4-5} \cmidrule(lr){6-7}
 & w/ cogmap & w/o cogmap & w/ cogmap & w/o cogmap & w/ cogmap & w/o cogmap \\
\midrule
GPT-4o~\cite{openai2023gpt4} & $0.696$ & $0.645$ & $0.342$ & $0.294$ & $\textbf{0.216}$ & $\textbf{0.285}$ \\
Gemini-1.5-Pro~\cite{geminiteam2024gemini15unlockingmultimodal} & $\textbf{0.744}$ & $\textbf{0.711}$ & $\textbf{0.417}$ & $\textbf{0.424}$ & $0.074$ & $0.109$\\
Llama-3.2-90B~\cite{grattafiori2024llama3herdmodels} & $0.697$ & $\underline{0.674}$ & $\underline{0.368}$ & $0.323$ & $\underline{0.196}$ & $0.123$ \\
Qwen2.5-VL-72B~\cite{bai2023qwenvlversatilevisionlanguagemodel} & $0.558$ & $0.498$ & $0.299$ & $0.325$ & $0.156$ & $\underline{0.181}$ \\
InternVL2\_5-78B~\cite{chen2025expandingperformanceboundariesopensource} & $0.683$ & $0.616$ & $0.365$ & $\underline{0.397}$ & $0.059$ & $0.087$ \\
LLaVA-v1.5-13B~\cite{liu2024improvedbaselinesvisualinstruction} & $0.598$ & $0.553$ & $0.245$ & $0.226$ & $0.144$ & $0.169$ \\
deepseek-v12~\cite{lu2024deepseekvlrealworldvisionlanguageunderstanding} & $\underline{0.734}$ & $0.579$ & $0.091$ & $0.055$ & $0.129$ & $0.131$ \\
Janus-Pro-7B~\cite{chen2025janusprounifiedmultimodalunderstanding} & $0.487$ & $0.405$ & $0.200$ & $0.159$ & $0.068$ & $0.159$ \\
\bottomrule
\end{tabular}%
}
\vspace{-16pt}
\end{table}

\noindent \textbf{Impact of Cognitive Maps on Spatial Reasoning.}
Table~\ref{tab:comparison_withcogmap} reveals interesting patterns regarding the utility of cognitive maps for spatial reasoning tasks. For object counting, all models benefit from having a cognitive map, with improvements ranging from $2\%$ to $15\%$. However, the impact on relative distance reasoning is mixed: some models (notably GPT-4o, Llama-3.2-90B, and deepseek-v12) improve substantially with cognitive maps, while others (Gemini-1.5-Pro, Qwen2.5-VL-72B) perform slightly better without them. Most surprisingly, for relative direction tasks, seven out of eight models perform better without cognitive maps, with only Llama-3.2-90B showing improvement when using cognitive maps. This counterintuitive finding suggests that the imperfect cognitive maps generated by these models may actually interfere with their innate ability to reason about directional relationships in panoramic images. The cognitive overhead of reconciling potentially inaccurate map representations with the original visual information appears to degrade performance on orientation-sensitive tasks, while still providing benefits for simpler counting operations. The qualitative examples of omnidirectional spatial reasoning are shown in Figure~\ref{case} and more case are shown in Appendix~\ref{app:more_cases}.

\begin{figure}[t!]
\centering
\includegraphics[width=\textwidth]{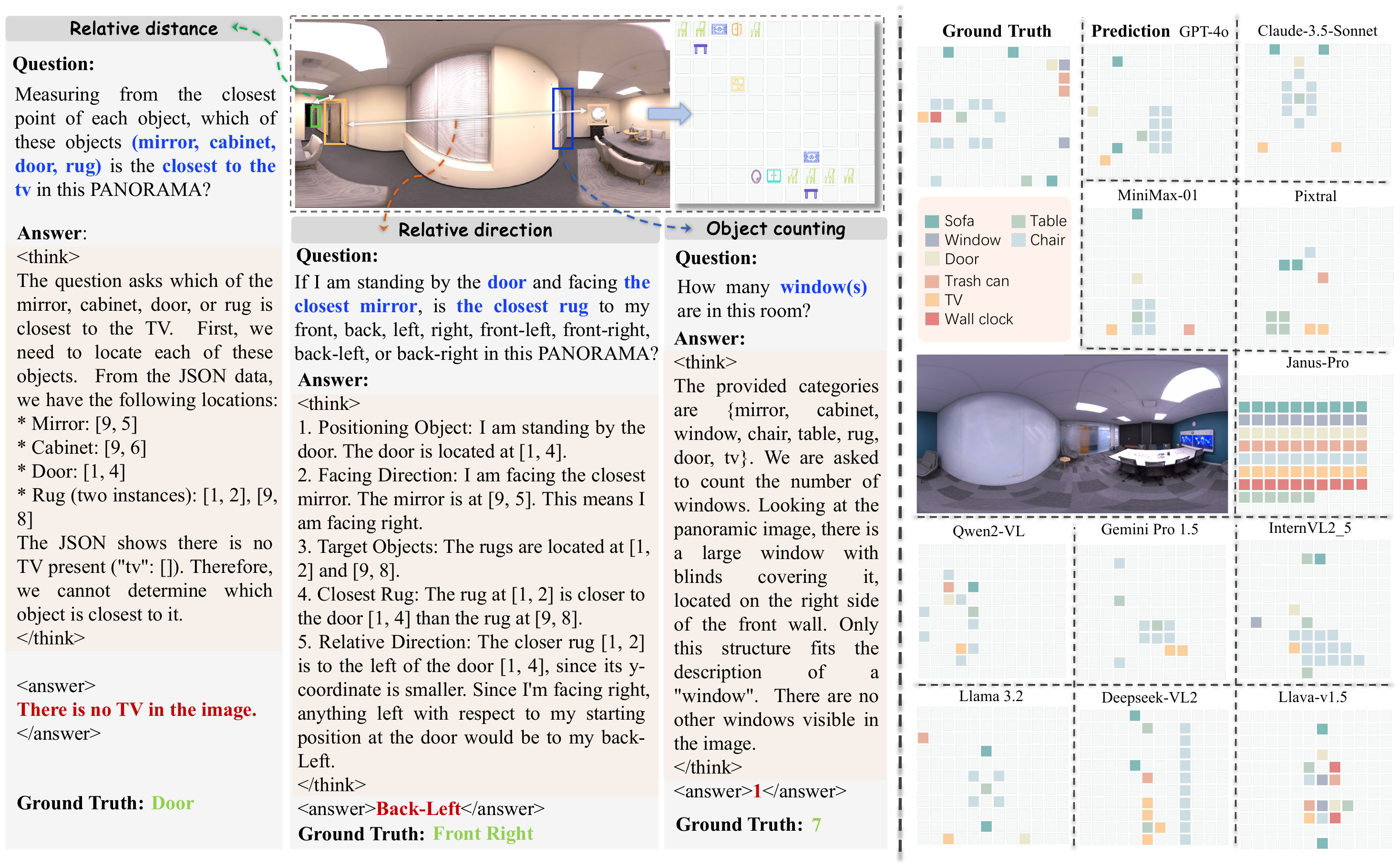}
\caption{Qualitative examples of omnidirectional spatial reasoning on our OSR-dataset.}
\label{case}
\vspace{-12pt}
\end{figure}
\begin{table}[t!]
\caption{\textbf{Model Performance with Negative Sampling.} Metrics include precision, recall, hallucination scores, and spatial reasoning.}
\label{tab:negative_sampling}
\centering
\renewcommand{\tabcolsep}{2.5pt}
\resizebox{\textwidth}{!}{%
\begin{tabular}{l cccccccc}
\toprule
\textbf{Model} & \textbf{Precision} & \textbf{Recall} & \textbf{CHAIR\_S} & \textbf{CHAIR\_I} & \textbf{CHAIR\_ins} & \textbf{obj count} & \textbf{rel distance} & \textbf{rel direction} \\
\midrule
GPT-4o~\cite{openai2023gpt4} & $\textbf{0.303}$ & $0.197$ & $0.791$ & $\textbf{0.198}$ & $\textbf{0.180}$ & $0.589$ & $\underline{0.435}$ & $0.137$ \\
Gemini-1.5-Pro~\cite{geminiteam2024gemini15unlockingmultimodal} & $0.269$ & $\textbf{0.211}$ & $0.852$ & $\underline{0.202}$ & $\textbf{0.180}$ & $\underline{0.602}$ & $\textbf{0.441}$ & $\underline{0.230}$ \\
Llama-3.2-90B~\cite{grattafiori2024llama3herdmodels} & $0.238$ & $0.201$ & $0.875$ & $0.339$ & $0.339$ & $0.535$ & $0.414$ & $0.140$ \\
Qwen2.5-VL-72B~\cite{bai2023qwenvlversatilevisionlanguagemodel} & $\textbf{0.303}$ & $0.189$ & $\textbf{0.767}$ & $0.314$ & $0.306$ & $0.579$ & $0.319$ & $\textbf{0.232}$ \\
InternVL2\_5-78B~\cite{chen2025expandingperformanceboundariesopensource} & $\underline{0.278}$ & $\underline{0.210}$ & $\underline{0.782}$ & $0.248$ & $\underline{0.243}$ & $0.567$ & $0.378$ & $0.156$ \\
LLaVA-v1.5-13B~\cite{liu2024improvedbaselinesvisualinstruction} & $0.083$ & $0.062$ & $0.884$ & $0.291$ & $0.284$ & $0.523$ & $0.175$ & $0.103$ \\
deepseek-vl2~\cite{lu2024deepseekvlrealworldvisionlanguageunderstanding} & $0.095$ & $0.124$ & $0.844$ & $0.380$ & $0.373$ & $\textbf{0.641}$ & $0.064$ & $0.047$ \\
Janus-Pro-7B~\cite{chen2025janusprounifiedmultimodalunderstanding} & $0.001$ & $0.002$ & $0.872$ & $0.363$ & $0.277$ & $0.510$ & $0.160$ & $0.004$ \\
\bottomrule
\end{tabular}%
}
\vspace{-20pt}
\end{table}
\begin{wrapfigure}{r}{0.35\textwidth}
\centering
\vspace{-5pt}
\captionof{table}{Success rates of different models in constructing properly formatted cognitive maps.}
\label{tab:cogmap_success_rate}
\renewcommand{\tabcolsep}{1pt}
\resizebox{\linewidth}{!}{
\begin{tabular}{lc}
\toprule
\textbf{Model} & \textbf{Success Rate (\%)} \\
\midrule
Gemini-1.5-Pro~\cite{geminiteam2024gemini15unlockingmultimodal} & $99.50$ \\
GPT-4o~\cite{openai2023gpt4} & $97.50$ \\
Qwen2.5-VL-72B~\cite{bai2023qwenvlversatilevisionlanguagemodel} & $95.60$ \\
InternVL2\_5-78B~\cite{chen2025expandingperformanceboundariesopensource} & $92.40$ \\
Llama-3.2-90B~\cite{grattafiori2024llama3herdmodels} & $79.15$ \\
deepseek-vl2~\cite{lu2024deepseekvlrealworldvisionlanguageunderstanding} & $44.83$ \\
LLaVA-v1.5-13B~\cite{liu2024improvedbaselinesvisualinstruction} & $10.84$ \\
Janus-Pro-7B~\cite{chen2025janusprounifiedmultimodalunderstanding} & $4.29$ \\
\bottomrule
\end{tabular}}
\vspace{-10pt}
\end{wrapfigure}
\noindent \textbf{Hallucination Robustness.}
Under the negative sampling condition (Table~\ref{tab:negative_sampling}), all models show substantial hallucination tendencies, with CHAIR\_S scores ranging from $0.767$ to $0.884$ (lower is better). Proprietary models are more robust, with GPT-4o achieving the lowest CHAIR\_I ($0.198$) and CHAIR\_instance ($0.180$) scores, indicating fewer hallucinated object classes and instances. Among open-source models, Qwen2.5-VL-72B shows strong resistance, with the lowest CHAIR\_S score of $0.767$. All models experience performance degradation in spatial reasoning tasks when negative distractor objects are introduced, highlighting limited robustness to interference. Interestingly, hallucination resistance does not perfectly correlate with spatial reasoning performance—deepseek-vl2 achieves the highest object counting score ($0.641$) despite relatively high hallucination metrics. This discrepancy likely results from deepseek-vl2's architecture, which includes specialized tokens for object localization and reasoning, allowing strong counting capabilities despite weaker hallucination resistance.

\noindent \textbf{Instruction Following and Format Adherence}
The vast differences in instruction-following capabilities across models are quantitatively demonstrated in Table~\ref{tab:cogmap_success_rate}, which shows the success rates in generating properly formatted cognitive maps. Proprietary models excel at this task, with Gemini-1.5-Pro and GPT-4o achieving success rates of $99.50$\% and $97.50$\% respectively. Among open-source models, larger models like Qwen2.5-VL-72B ($95.60$\%) and InternVL2\_5-78B ($92.40$\%) approach proprietary-level performance, while smaller models struggle dramatically. The success rate drops to just $10.84$\% for LLaVA-v1.5-13B and a mere $4.29$\% for Janus-Pro-7B. This strong correlation between parameter count and instruction following capability explains much of the performance gap observed in Table~\ref{tab:omni-cogmap_Generation_comparison}. Models with low success rates in generating properly formatted cognitive maps inevitably achieve poor F1 scores, as their failures stem not just from limitations in visual-spatial understanding but from fundamental difficulties in following structured response protocols for complex spatial representations.
\begin{wraptable}{r}{8cm}
\vspace{-8pt}
\caption{\textbf{Thinking-mode Prompt Result.} Performance improvements for Gemini-1.5-Pro using thinking-mode prompt (with and without Omni-cognitive map) on the non-negatively sampling version of the dataset.} 
\label{tab:cogmap_comparison}
\centering
\renewcommand{\tabcolsep}{1pt}
\resizebox{\linewidth}{!}{
\begin{tabular}{lccc}
\toprule
& \textbf{object count} & \textbf{rel distance} & \textbf{rel direction} \\
\midrule
w/ omni-cogmap & $0.745$~\textcolor{green}{$(+0.001)$} & $0.424$~\textcolor{green}{$(+0.007)$} & $0.135$~\textcolor{green}{$(+0.061)$} \\
w/o omni-cogmap & $0.748$~\textcolor{green}{$(+0.037)$} & $0.434$~\textcolor{green}{$(+0.009)$} & $0.092$~\textcolor{red}{$(-0.018)$} \\
\bottomrule
\end{tabular}}
\vspace{-12pt}
\end{wraptable}
\noindent \textbf{Enhanced Reasoning Through CoT Prompt.}
The results in Table~\ref{tab:cogmap_comparison} show that prompting models to use structured reasoning improves performance across most tasks. For example, Gemini-1.5-Pro gains $+0.037$ in object counting without cognitive maps when guided to think step-by-step (templates in Appendix~\ref{prop:post_prompt}). This suggests that explicit reasoning scaffolds are particularly helpful when models rely solely on internal scene representations.  
However, gains are generally smaller when cognitive maps are provided, indicating that external spatial representations may overlap with structured reasoning. An exception is relative direction reasoning, which improves significantly ($+0.061$) with thinking-mode prompts, suggesting that explicit reasoning helps models better integrate directional information from cognitive maps. Overall, structured thinking appears to enhance the effective use of cognitive maps, reducing potential confusion and improving spatial understanding. These findings are further supported by the qualitative results presented in Figure~\ref{case}.

\section{Conclusion}  
\label{sec:conclusion}
In this paper, we introduced \textbf{OSR-Dataset} and \textbf{OSR-Bench}, the first benchmarks specifically designed to evaluate the spatial reasoning capabilities of Multimodal Large Language Models (MLLMs) in 360{\textdegree} panoramic environments. Our benchmark includes over $153,000$ question-answer pairs across three core spatial reasoning tasks—object counting, relative distance, and relative direction—grounded in high-fidelity 3D scene layouts. To improve evaluation rigor, we proposed a negative sampling strategy to systematically test for hallucination and grounding robustness.  
We also developed a two-stage evaluation framework that combines rotation-invariant cognitive map alignment with both rule-based and LLM-based question answering, providing fine-grained, interpretable metrics. Our evaluation of eight state-of-the-art MLLMs revealed significant gaps in their spatial reasoning performance under panoramic settings, underscoring the need for more perceptually grounded architectures.  

\noindent \textbf{Limitations \& Broader Impacts}
While OSR-Bench provides a comprehensive evaluation framework for omnidirectional spatial reasoning, several limitations remain. The benchmark primarily focuses on indoor environments, which may not fully capture the complexities of outdoor scenes, such as dynamic objects, varying weather conditions, and large-scale spatial structures. Our work advances perceptually grounded MLLMs, with applications in autonomous navigation, augmented reality, and robotics. However, the improved spatial reasoning enabled by this research may raise privacy and surveillance concerns, as panoramic cameras are commonly used in security systems. There is also a risk of misuse in military or surveillance contexts. Researchers should consider these ethical implications when deploying MLLMs trained on spatial reasoning benchmarks.

\noindent \textbf{Maintenance Plan and Future Work}
To ensure ease of access and future maintenance, the dataset and models evaluated in our paper are hosted on websites and cloud-based storage platforms. And our future work will focus on: \textbf{(I)} Extending to real-world scenes beyond synthetic indoor environments. \textbf{(II)} Evaluating model generalization to out-of-distribution scenarios, including unfamiliar objects and novel scenes. \textbf{(III)} Developing more sophisticated reasoning tasks that capture dynamic scene understanding. \textbf{(IV)} Exploring post-training strategies to enhance panoramic spatial reasoning abilities.
OSR-Bench provides a foundation for advancing omnidirectional perception research, and we are committed to expanding this resource for the community.

    \clearpage
    {\small
    \bibliographystyle{unsrtnat} 
    \bibliography{reference}
    }
    \clearpage


\appendix


\section{Algorithm Detail}

In this section, we provide more details on the construction pipeline for our \textit{OSR-Bench}.

\subsection{QA Pair Generation}
\label{alg:QAGeneration}

Algorithm~\ref{alg:QA_generation} details our systematic approach to generating question-answer pairs from omni-cognitive maps. For each scene, we generate three types of questions (object counting, relative distance, and relative direction) using structured templates. The algorithm takes as input an omni-cognitive map, a class count dictionary, and the expected  max number of questions per type. We determine the expected maximum number of questions based on the similarity of the scenes to which the images correspond. For images obtained from DeepPanoContext~\cite{zhang2021deeppanocontext}, we set the expected number of questions to 10. For images obtained from ReplicaPano~\cite{dong2024panocontext}, we set the expected number of questions to 5.

To eliminate the duplication and ambiguity of the problem, we sample the objects in the scene without replacement and ensure the uniqueness of the positioning objects when generating non-negative sampling problems for relative distance and relative direction. In particular, if there are less than 5 object categories in the scene, we will not generate the above two types of problems. At the same time, we ensure that when generating relative direction problems, the three types of objects in the problem do not appear in the same cognitive map grid unit.

\begin{algorithm}
\caption{QA Pair Generation}
\label{alg:QA_generation}
\begin{algorithmic}[1]
\Require Omni-cognitve map $C$, Class count dictionary $\mathcal{D}$, Questions per type $N$
\State \textbf{Define Question Templates:}
\State $T_{count} \leftarrow$ ``How many ${object}$(s) are in this room?''
\State $T_{dist} \leftarrow$ ``Measuring from the closest point of each object, which of these objects (${candidates}$) is the closest to the ${positioning}$ in this PANORAMA?''
\State $T_{dir} \leftarrow$ ``If I am standing by the ${positioning}$ and facing the closest ${orienting}$, is the closest ${querying}$ to my front, back, left, right, front-left, front-right, back-left, or back-right in this PANORAMA?''

\State $\mathcal{O} \leftarrow$ all object classes in $\mathcal{D}$
\State $\mathcal{O}_{unique} \leftarrow {o \in \mathcal{O} \mid \mathcal{D}[o] = 1}$
\State $\mathcal{Q} \leftarrow \emptyset$

\State // Generate object counting questions
\State Sample $\min(N, |\mathcal{O}|)$ objects from $\mathcal{O}$ without replacement
\For{each sampled object $o$}
\State $q \leftarrow T_{count}$ with ${object}$ replaced by $o$
\State $a \leftarrow \mathcal{D}[o]$ //answer for $o$
\State Add to $\mathcal{Q}$
\EndFor

\State // Generate relative distance questions
\For{$i = 1$ to $N$}
\State Sample positioning object $p$ from $\mathcal{O}_{unique}$
\State Sample distinct candidates ${c_1,...,c_4}$ from $\mathcal{O} \setminus {p}$
\State $q \leftarrow T_{dist}$ with ${positioning}$ replaced by $p$, ${candidates}$ replaced by ${c_1,...,c_4}$
\State $a \leftarrow c_i$ find by Euclidean distance in grid coordinates
\State Add to $\mathcal{Q}$
\EndFor

\State // Generate relative direction questions
\For{$i = 1$ to $N$}
\State Sample positioning object $p$ from $\mathcal{O}_{unique}$
\State Sample distinct orienting object $o \in \mathcal{O} \setminus {p}$ and querying object $q \in \mathcal{O} \setminus {p, o}$
\State $q \leftarrow T_{dir}$ with ${positioning}$ replaced by $p$, ${orienting}$ replaced by $o$, and ${querying}$ replaced by $q$
\State $a \leftarrow $Direction Computed based on vector relationships in grid coordinates
\State Add to $\mathcal{Q}$
\EndFor
\State \Return $\mathcal{Q}$
\end{algorithmic}
\end{algorithm}

\subsection{Negative Sampling for Hallucination Testing}
\label{alg:negative-sampling}

\begin{figure}
\centering
\includegraphics[width=\textwidth]{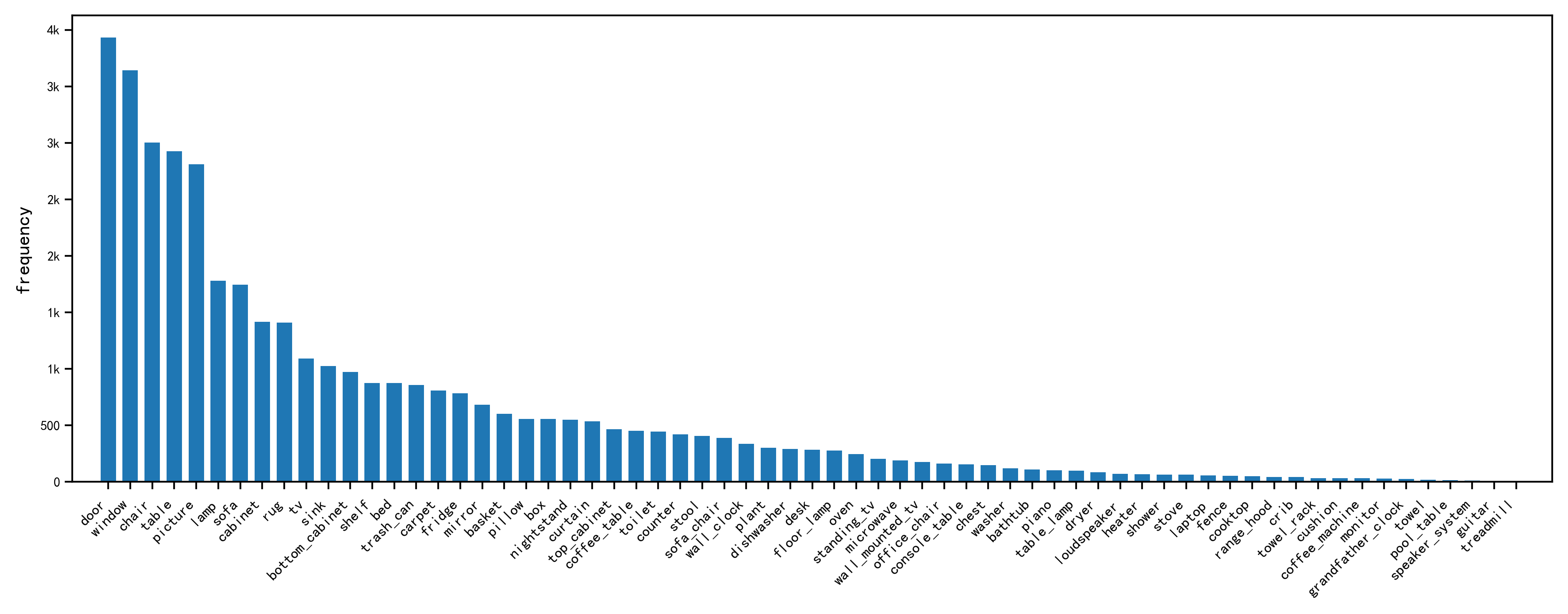}
\caption{Object frequency of the entire dataset.}
\label{fig:object_frequency}
\end{figure}

Algorithm~\ref{alg:negative_sampling_detail} outlines our negative sampling strategy for evaluating model robustness against hallucination. We first compute global statistics on object frequency (as shown in Figure~\ref{fig:object_frequency}) and co-occurrence patterns across the dataset. For each scene, we identify two types of distractor objects: \textcircled{1}  frequently occurring objects that are absent from the current scene ($\mathcal{M}_f$) and \textcircled{2} objects that commonly co-occur with present objects but are missing from the current scene ($\mathcal{M}_c$). These negative distractor objects are incorporated into the object pool used for Omni-Cognitive map generation prompt and QA generation. Specifically, we set the number of $\mathcal{M}_f$ and $\mathcal{M}_c$ for each image to 5 when performing negative sampling.

\begin{algorithm}
\caption{Negative Sampling for Hallucination Testing}
\label{alg:negative_sampling_detail}
\begin{algorithmic}[1]
\Require Omni-Cognitive map $M$, Parameters $n$, $m$
\State Compute global object frequency statistics $\mathcal{F}$ and global co-occurrence statistics $\mathcal{C}$
\State Extract present objects $\mathcal{P}$ from omni-cognitive map $M$

\State $\mathcal{M}_f \leftarrow$ Top $n$ most frequent objects from $\mathcal{F}$ that are not in $\mathcal{P}$
\State $\mathcal{M}_c \leftarrow \emptyset$ //Co-occurring but missing objects
\State $candidates \leftarrow \emptyset$
\For{each object $p \in \mathcal{P}$}
\For{each object $o$ co-occurring with $p$, sorted by co-occurrence frequency}
\If{$o \notin \mathcal{P}$ and $o \notin \mathcal{M}_f$ and $o \notin candidates$}
\State $candidates \leftarrow candidates \cup {o}$
\If{$|candidates| = m$}
\State \textbf{exit} loop
\EndIf
\EndIf
\EndFor
\EndFor
\State $\mathcal{M}_c \leftarrow$ Top $m$ objects from $candidates$
\State $\mathcal{O}_{aug} \leftarrow \mathcal{P} \cup \mathcal{M}_f \cup \mathcal{M}_c$ //Augmented object set
\State Generate QA pairs using $\mathcal{O}_{aug}$ with rules:
\State \quad - For counting: Return 0 for objects in $\mathcal{M}_f \cup \mathcal{M}_c$
\State \quad - For spatial questions: Return appropriate ``not found'' responses when needed
\State \textbf{return} QA pairs with both real and hallucinated objects
\end{algorithmic}
\end{algorithm}

\section{Dataset Detail}
\label{app:data_detail}

This section provides statistical insights into the composition and characteristics of our OSR-Bench dataset.

Our dataset contains a large number of question-answer pairs across different reasoning categories. For the standard dataset, we have a total of $74,326$ questions distributed as follows:
\begin{itemize}
    \item Relative distance questions: $26,240$
    \item Object counting questions: $24,988$
    \item Relative direction questions: $23,098$
\end{itemize}

For the dataset with negative sampling , we have $78,986$ questions distributed as:
\begin{itemize}
    \item Object counting questions: $27,125$
    \item Relative distance questions: $26,250$
    \item Relative direction questions: $25,611$
\end{itemize}

\begin{figure}
    \centering
    \begin{subfigure}[t]{0.5\textwidth}
        \centering
        \adjustbox{max height=6cm, max width=\textwidth}{%
            \includegraphics{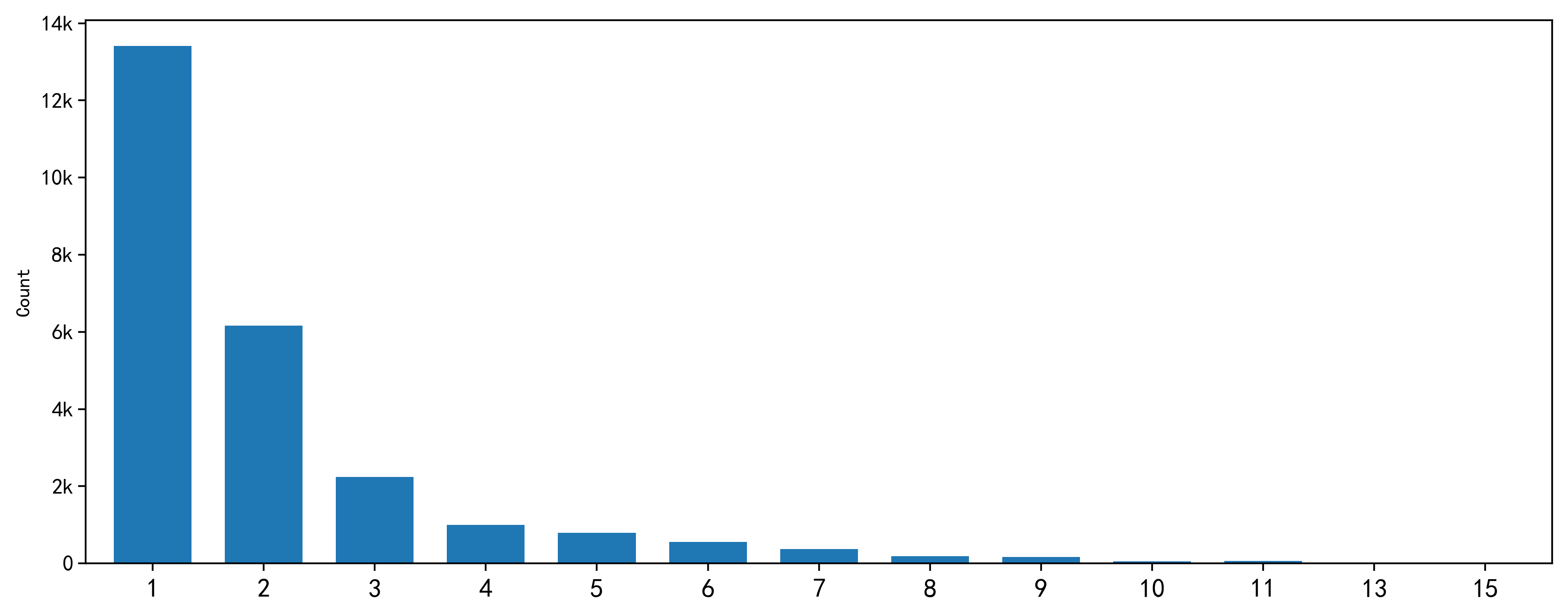}%
        }
        \caption{QA pairs without negative sampling.}
    \end{subfigure}%
    \hfill
    \begin{subfigure}[t]{0.5\textwidth}
        \centering
        \adjustbox{max height=6cm, max width=\textwidth}{%
            \includegraphics{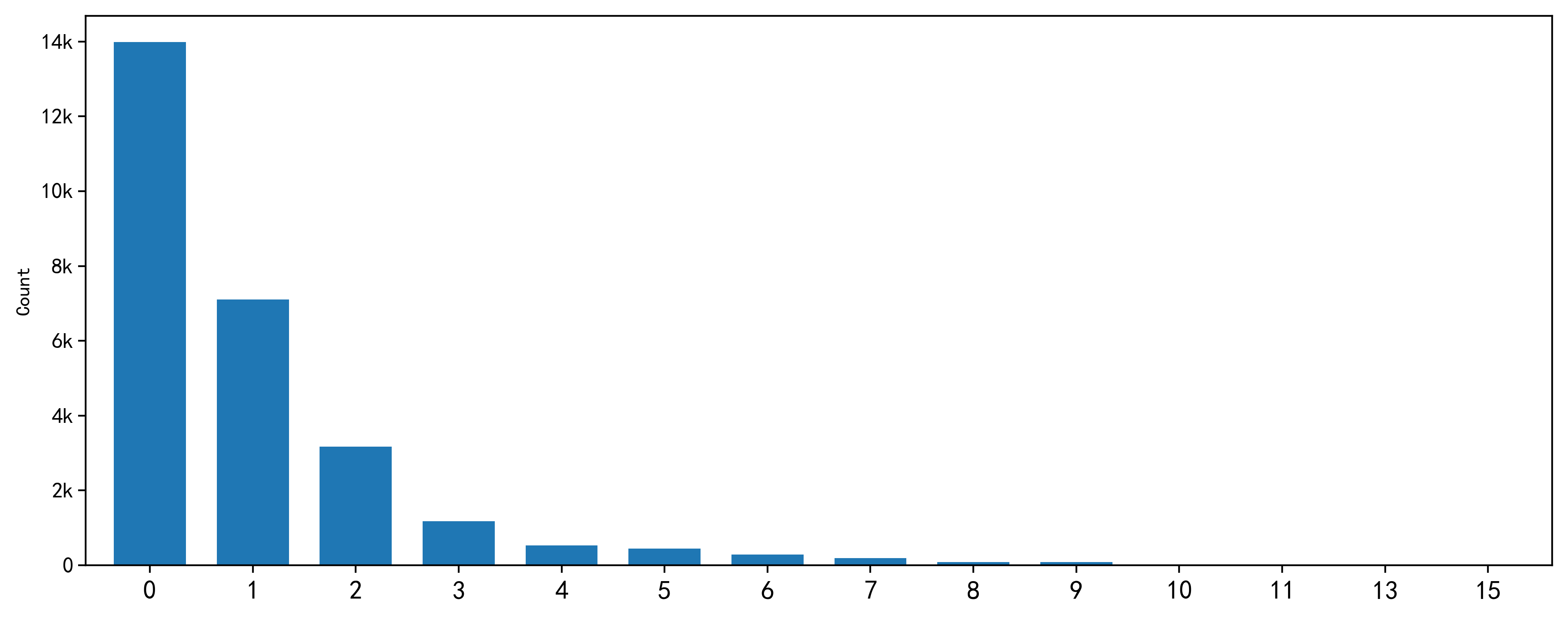}%
        }
        \caption{QA pairs with negative sampling.}
    \end{subfigure}
    \caption{Answer distribution of Object Counting question.}
    \label{fig:object_counting}
\end{figure}

\begin{figure}
    \centering
    \subfloat[QA pairs without negative sampling.]{\includegraphics[width=\textwidth]
    {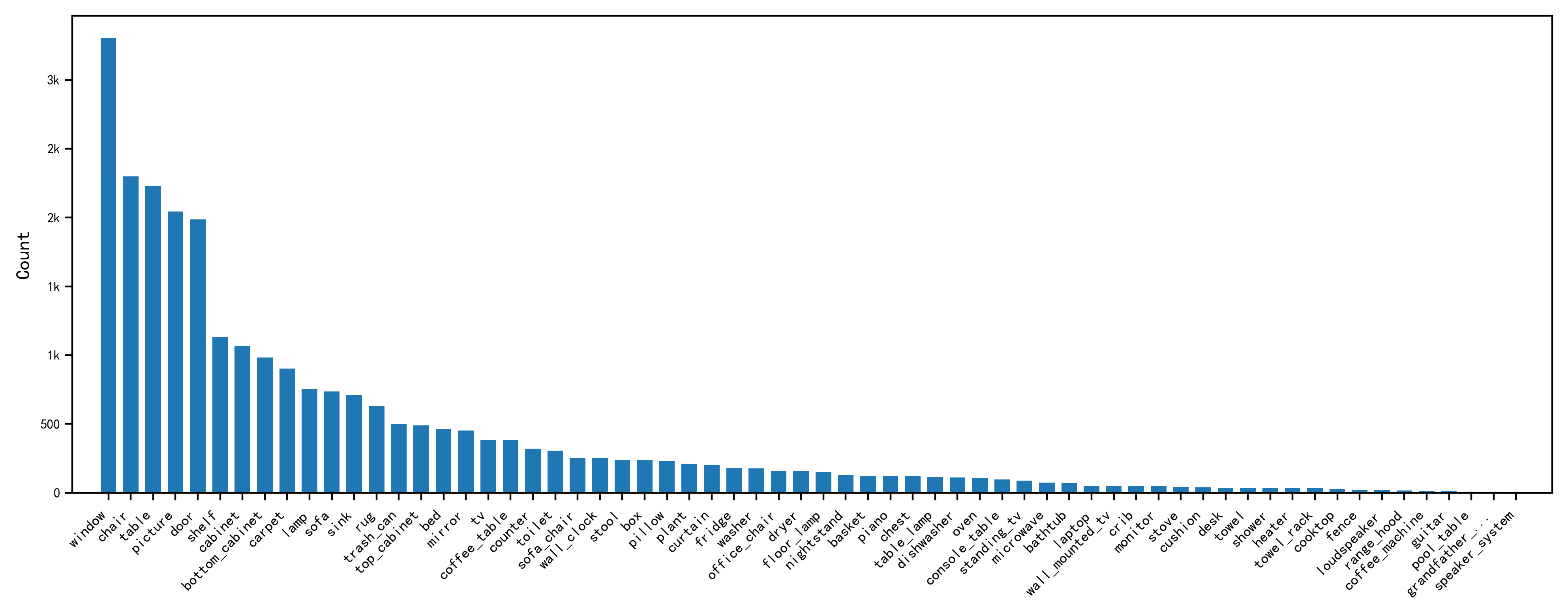}}
    
    \subfloat[QA pairs with negative sampling.]
    {\includegraphics[width=\textwidth]{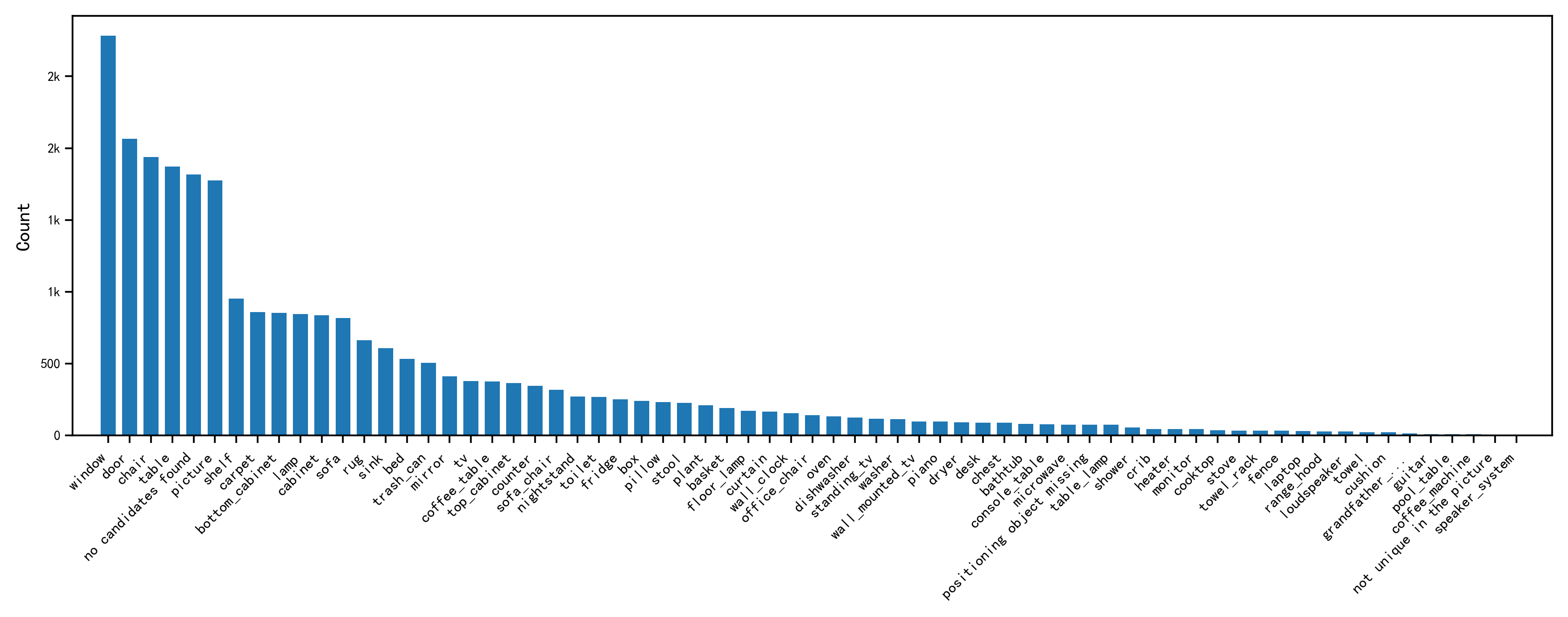}}
    \caption{Answer distribution of Relative Distance question.}
    \label{fig:relative_distance}
\end{figure}

\begin{figure}
    \centering
    \begin{subfigure}[t]{0.5\textwidth}
        \centering
        \adjustbox{max height=6cm, max width=\textwidth}{%
            \includegraphics{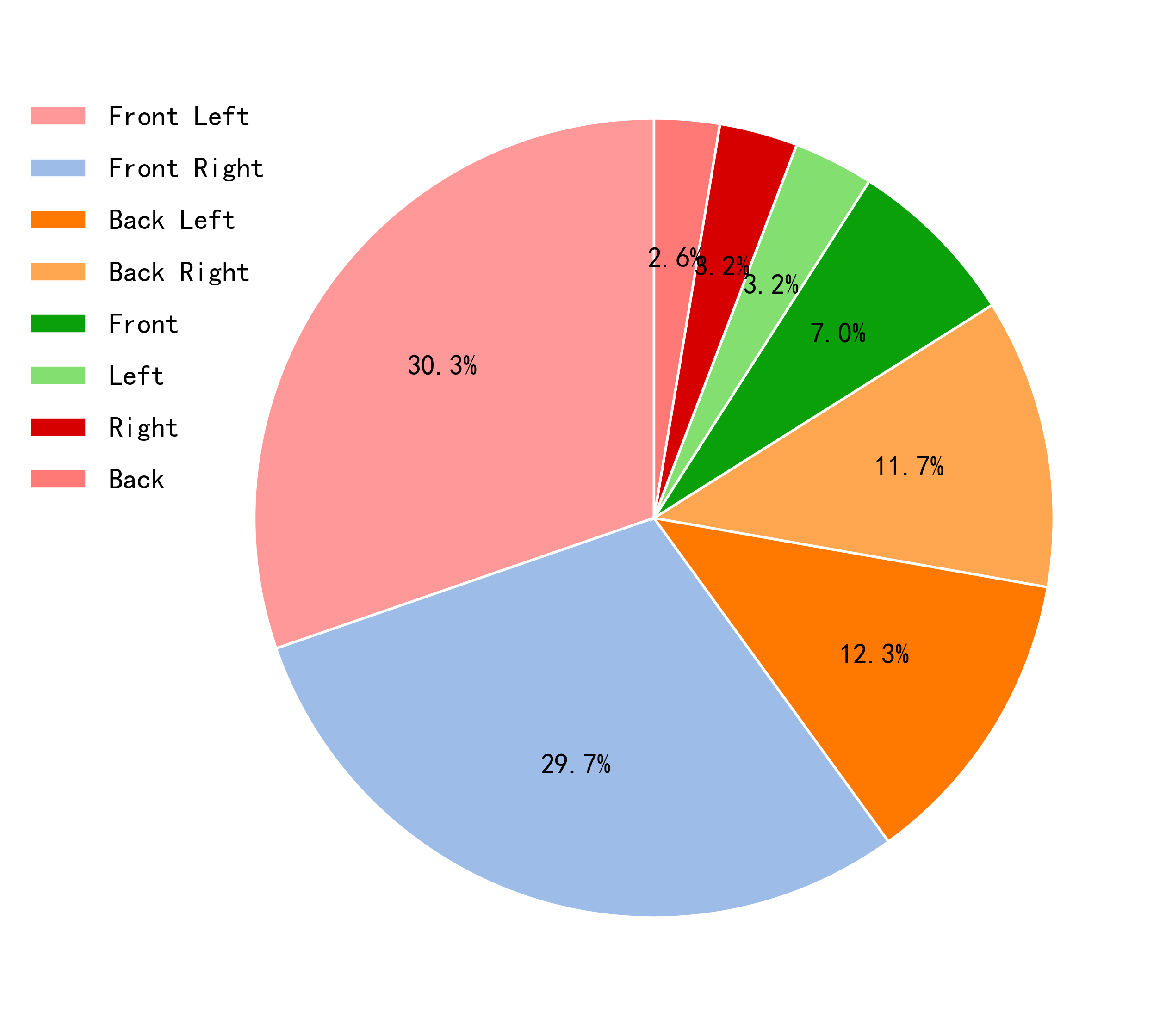}%
        }
        \caption{QA pairs without negative sampling.}
    \end{subfigure}%
    \hfill
    \begin{subfigure}[t]{0.5\textwidth}
        \centering
        \adjustbox{max height=6cm, max width=\textwidth}{%
            \includegraphics{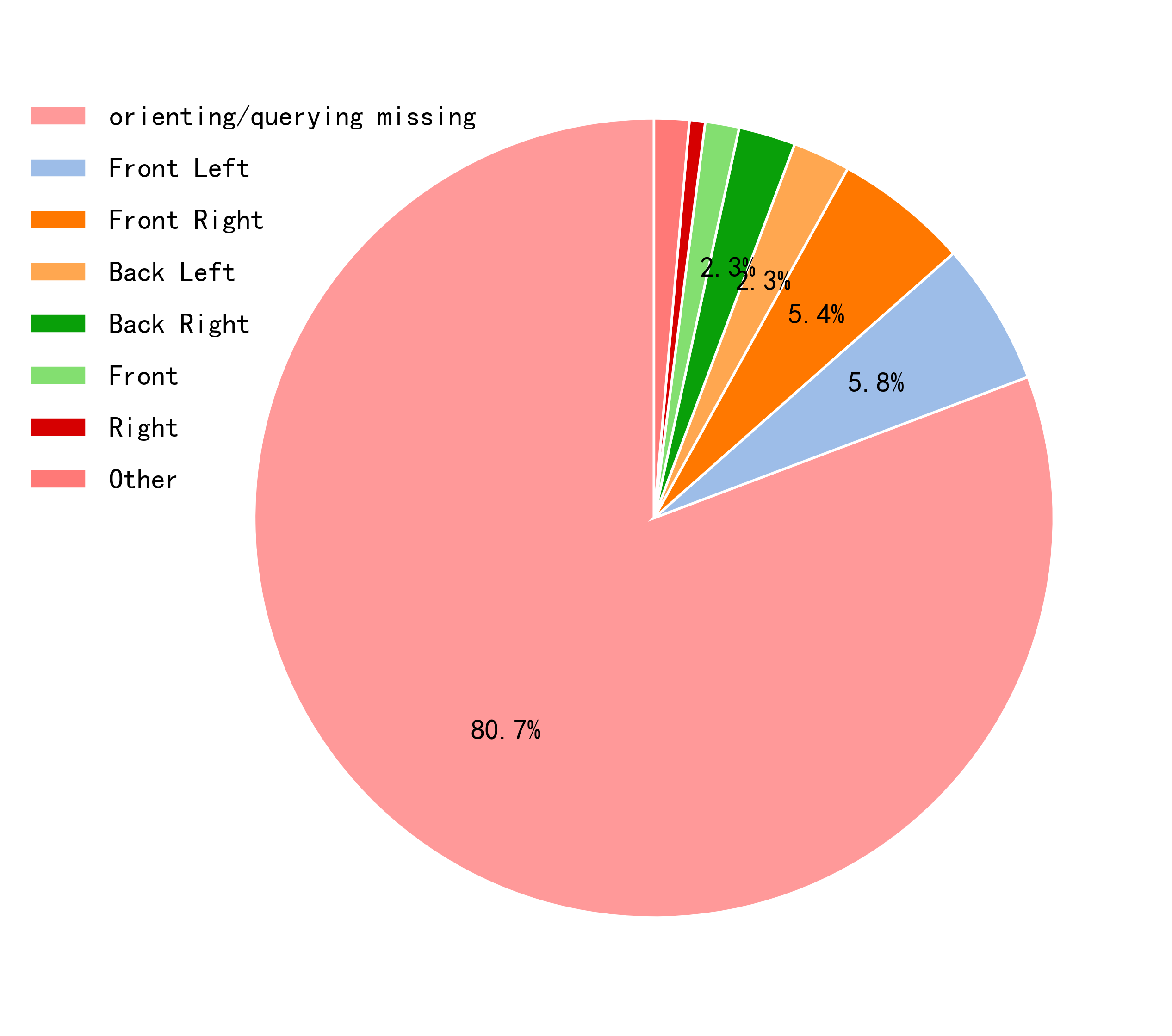}%
        }
        \caption{QA pairs with negative sampling.}
    \end{subfigure}
    \caption{Answer distribution of Relative Direction question.}
    \label{fig:relative_direction}
\end{figure}

Figures~\ref{fig:object_counting}-\ref{fig:relative_direction} present the answer distributions for each question type, both with and without negative sampling. 

\section{Prompt Template}
\definecolor{pastelBlue}{RGB}{210, 225, 245}
\definecolor{pastelYellow}{RGB}{250, 245, 210}
\definecolor{pastelPurple}{RGB}{230, 215, 240}
\definecolor{titleTextColor}{RGB}{75, 0, 130}

\newcommand{\highlight}[1]{\colorbox{pastelPurple}{#1}}

This section details the prompt templates used throughout our evaluation pipeline.

\subsection{Omni-Cognitive Map Generate Prompt}
\label{prop:cogmap_generate}

The cognitive map generation prompt instructs models to analyze a panoramic image and construct a structured spatial representation of the scene. In actual application, in the case of non-negative sampling, we fill the \{object\_pool\} with the object categories contained in the Ground truth. In the case of negative sampling, we add negative distractor objects to the \{object\_pool\}.

\begin{tcolorbox}[
  colback=pastelYellow,
  colframe=pastelBlue!70!black,
  colbacktitle=pastelBlue,
  coltitle=titleTextColor,
  fonttitle=\bfseries,
  title=Cognitive Map Prompt,
  rounded corners,
  boxrule=0.5mm,
  breakable,
  width=\textwidth
]
\textbf{[Task]} \\
This PANORAMA captures an indoor scene. Your objective is to identify specific objects within the panorama, understand the spatial arrangement of the scene, and estimate the center point of each object, assuming the entire scene is represented by a 10x10 grid. \\
\textbf{[Rule]} \\
1. We provide the categories to care about in this scene: \highlight{\{object\_pool\}}. Focus ONLY on these categories. \\
2. Estimate the center location of each instance within the provided categories, assuming the entire scene is represented by a 10x10 grid. \\
3. If a category contains multiple instances, include all of them. \\
4. Each object’s estimated location should accurately reflect its real position in the scene, preserving the relative spatial relationships among all objects. \\
5. Consider object distance from camera when positioning them. Closer objects should be placed near the center of the grid, while distant objects should be placed toward the edges. \\
6. If an object is partially visible or occluded, estimate its full position based on the visible parts. \\
\textbf{[Output]} \\
Present the estimated center locations for each object as a list within a dictionary. STRICTLY follow this JSON format And use INTEGERS within 10 to represent the coordinates: \{"category name": [(x\_1, y\_1), ...], ...\}
\end{tcolorbox}

\subsection{Pre and Post Prompt}
\label{prop:post_prompt}

When we do not prompt the model to generate cognitive maps but ask it to answer questions directly, we add a pre-prompt before the question. The input for the models is formatted as follows: \texttt{[Image Token][Pre-prompt][Question][Post-prompt]}

\begin{figure}[h!]
\centering
\includegraphics[width=\textwidth]{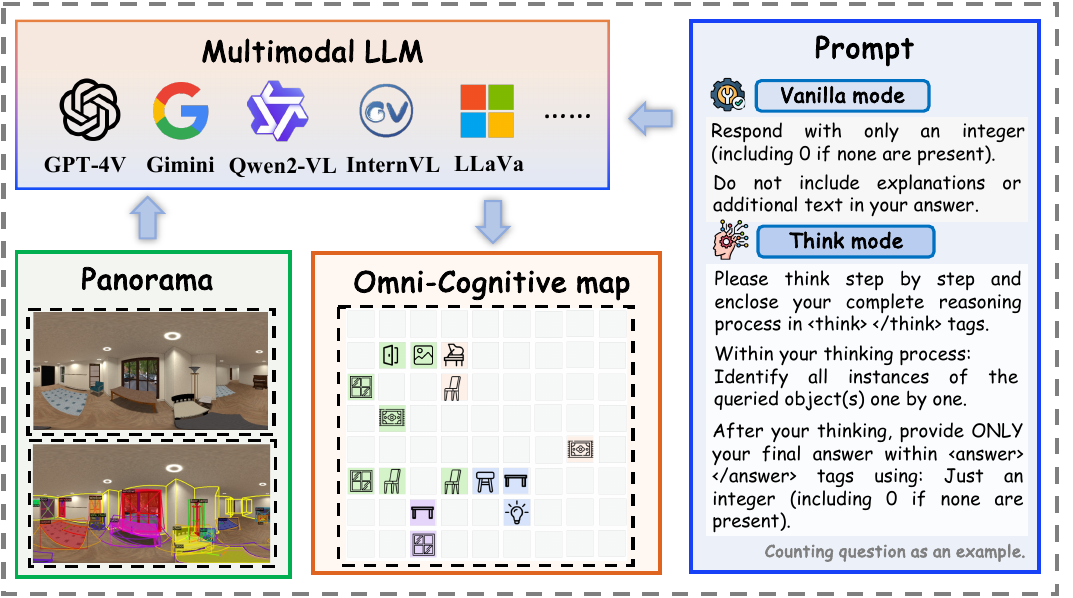}
\caption{Inference pipeline.}
\label{Inference pipeline}
\end{figure}

Figure~\ref{Inference pipeline} shows our complete reasoning process, which includes two evaluation modes: ``Vanilla Mode'' and ``Think Mode''. We switch between the two modes by using different post-prompts.

\begin{tcolorbox}[
  colback=pastelYellow,
  colframe=pastelBlue!70!black,
  colbacktitle=pastelBlue,
  coltitle=titleTextColor,
  fonttitle=\bfseries,
  title= Pre-Prompt,
  rounded corners,
  boxrule=0.5mm,
  breakable,
  width=\textwidth
]
According to the PANORAMA and the predicted objects' center locations, answer the following question:
\end{tcolorbox}

\subsection{LLM Evaluator Prompt}
\label{prop:llm_evalautor}

The LLM evaluator prompts enable automatic assessment of model responses, particularly for complex cases where rule-based evaluation may be insufficient. We provide three distinct evaluator prompts for the different question types. Each prompt includes specific evaluation rules and a structured JSON response format with both a binary score and confidence level. This approach ensures consistent assessment across diverse response formats from different models, especially for cases involving negative sampling where proper handling of non-existent objects is crucial. Specifically, we use DeepSeek-V3 as the evaluator, and replace \{question\},\{ground\_truth\}, \{model\_answer\} in the template with the real case.

\begin{tcolorbox}[
  colback=pastelYellow,
  colframe=pastelBlue!70!black,
  colbacktitle=pastelBlue,
  coltitle=titleTextColor,
  fonttitle=\bfseries,
  title= META\_PROMPT\_OBJECT\_COUNTING,
  rounded corners,
  boxrule=0.5mm,
  breakable,
  width=\textwidth
]
You are evaluating an object counting answer. \\
Question: \highlight{\{question\}} \\
Ground Truth: \highlight{\{ground\_truth\}} \\
Model Answer: \highlight{\{model\_answer\}} \\
Evaluation rules:\\
- The answer must be an integer (including 0) \\
- Correct if the model's integer exactly matches the ground truth \\
- Also correct if the model answer contains additional text but has the correct number \\
Please respond in JSON format: \\
\{ \\
    ``score'': 0 or 1,  \\
    ``confidence'': ``high'' or ``middle'' or ``low'' \\
\}
\end{tcolorbox}

\begin{tcolorbox}[
  colback=pastelYellow,
  colframe=pastelBlue!70!black,
  colbacktitle=pastelBlue,
  coltitle=titleTextColor,
  fonttitle=\bfseries,
  title= META\_PROMPT\_RELATIVE\_DISTANCE,
  rounded corners,
  boxrule=0.5mm,
  breakable,
  width=\textwidth
]
You are evaluating a relative distance answer. \\
Question: \highlight{\{question\}} \\
Ground Truth: \highlight{\{ground\_truth\}} \\
Model Answer: \highlight{\{model\_answer\}} \\
Evaluation rules:\\
- Normal answers are object names \\
- Special case answers may be one of these error messages: \\
    ``The positioning object is not found in the picture.'' \\
    ``The positioning object is not unique in the picture.'' \\
    ``None of the candidates were found in the picture.'' \\
- When the ground truth is one of these special error messages, the model answer must match exactly \\
- When the ground truth is an object name, the model answer is correct if it contains the matching object name (case insensitive) \\
Please respond in JSON format: \\
\{ \\
    ``score'': 0 or 1,  \\
    ``confidence'': ``high'' or ``middle'' or ``low'' \\
\}
\end{tcolorbox}

\begin{tcolorbox}[
  colback=pastelYellow,
  colframe=pastelBlue!70!black,
  colbacktitle=pastelBlue,
  coltitle=titleTextColor,
  fonttitle=\bfseries,
  title= META\_PROMPT\_RELATIVE\_DIRECTION,
  rounded corners,
  boxrule=0.5mm,
  breakable,
  width=\textwidth
]
You are evaluating a relative direction answer. \\
Question: \highlight{\{question\}} \\
Ground Truth: \highlight{\{ground\_truth\}} \\
Model Answer: \highlight{\{model\_answer\}} \\
Evaluation rules:\\
- Normal answers are direction words: ``Front'', ``Back'', ``Left'', ``Right'', ``Front-Left'', ``Front-Right'', ``Back-Left'', ``Back-Right'' \\
- Direction words are case insensitive, and hyphens are optional (\eg, ``front left'' equals ``Front-Left'') \\
- Special case answers may be one of these error messages: \\
    ``The positioning object is not found in the picture.'' \\
    ``The positioning object is not unique in the picture.'' \\
    ``The orienting object or querying object is not found in the picture.'' \\
- When the ground truth is one of these special error messages, the model answer must match exactly \\
Please respond in JSON format: \\
\{ \\
    ``score'': 0 or 1,  \\
    ``confidence'': ``high'' or ``middle'' or ``low'' \\
\}
\end{tcolorbox}

\section{Evaluation Details}
\label{evaluation_details}

To ensure reproducibility, we adopt a greedy decoding strategy for all models (\ie, the temperature is set to 0, and both top-p and top-k are set to 1 for proprietary models, do\_sample is set to \texttt{False} for open-source models).

A complete evaluation task (\ie, evaluating a specific model with or without a omni-cognitive map, in vanilla or think mode, on a non-negative or negative sampling dataset) requires approximately 4-5 days. For hardware requirements, models below 13B parameters use 1 NVIDIA A40 48G GPU, models between 13B and 30B use 3 NVIDIA A40 48G GPUs, while models exceeding 70B parameters require 3 NVIDIA A800 80G GPUs.

\begin{figure}[h!]
\centering
\includegraphics[width=0.5\textwidth]{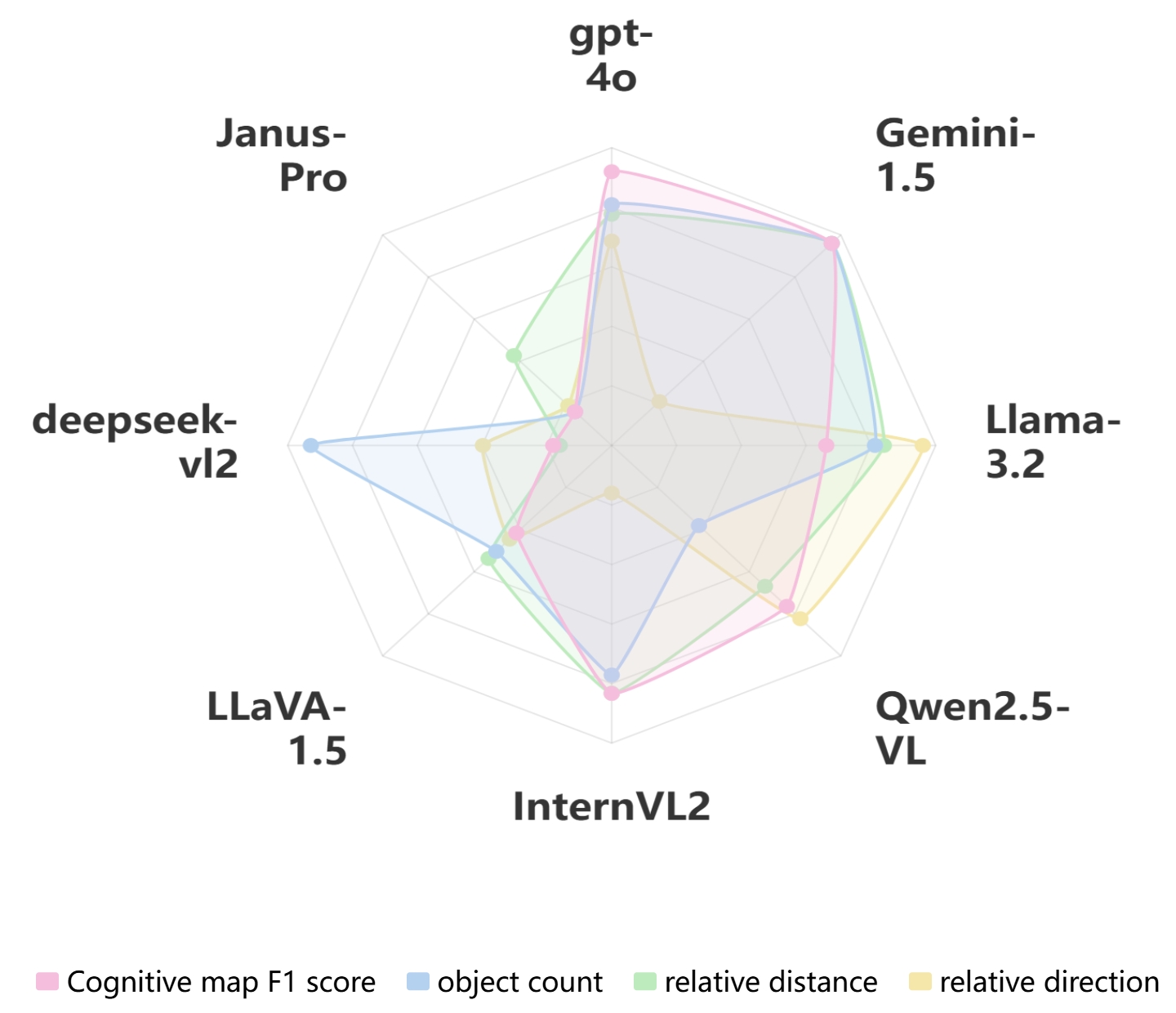}
\caption{\textbf{\textit{OSR}}-benchmark Overview}
\label{fig:radar_chart}
\end{figure}

\section{More Cases}
\label{app:more_cases}

\begin{figure}
\centering
\includegraphics[width=\textwidth]{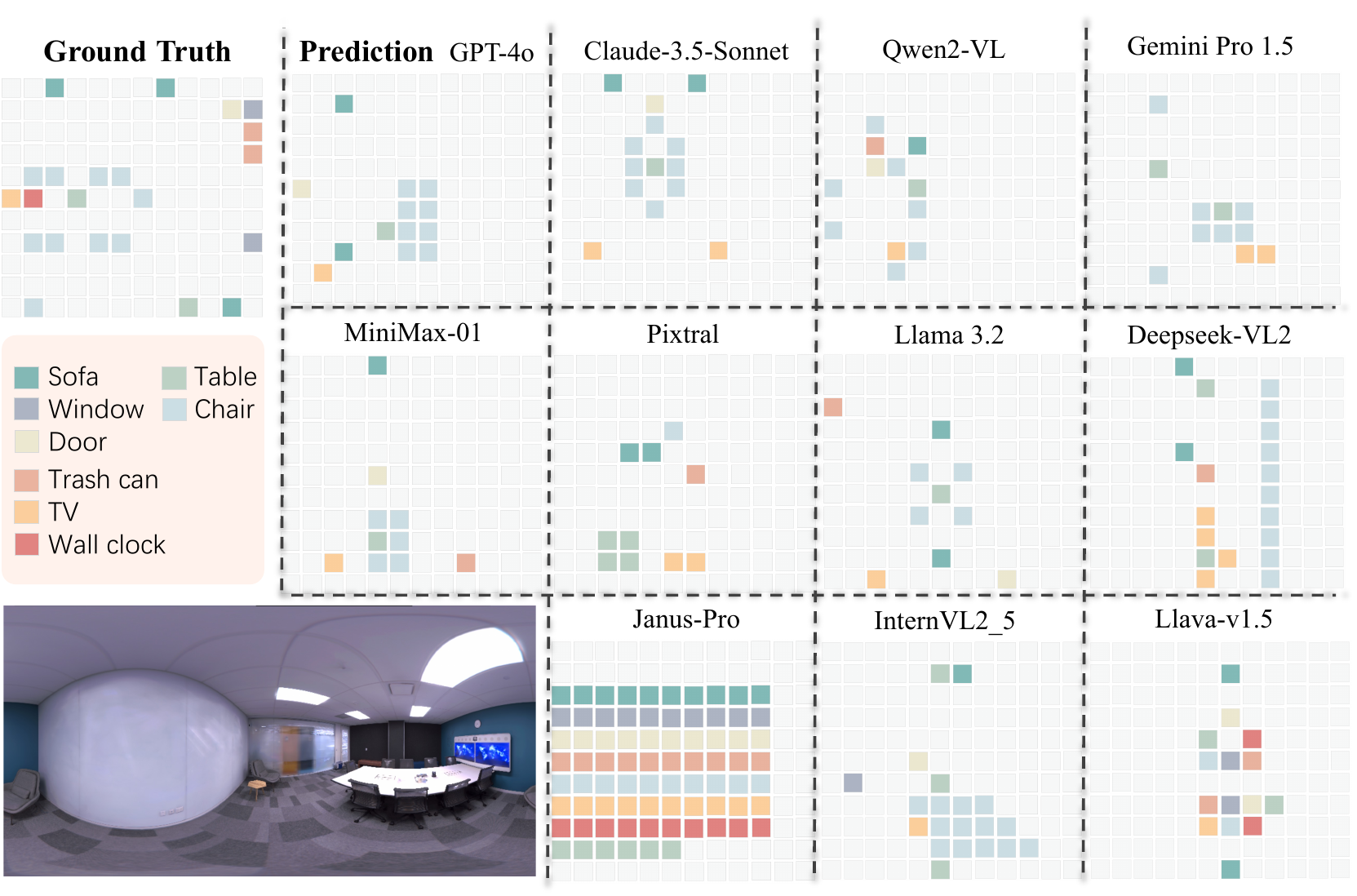}
\caption{Visualization of Omni-Cognitive map generation}
\label{cog_map}
\end{figure}

\begin{figure}
\centering
\includegraphics[width=\textwidth]{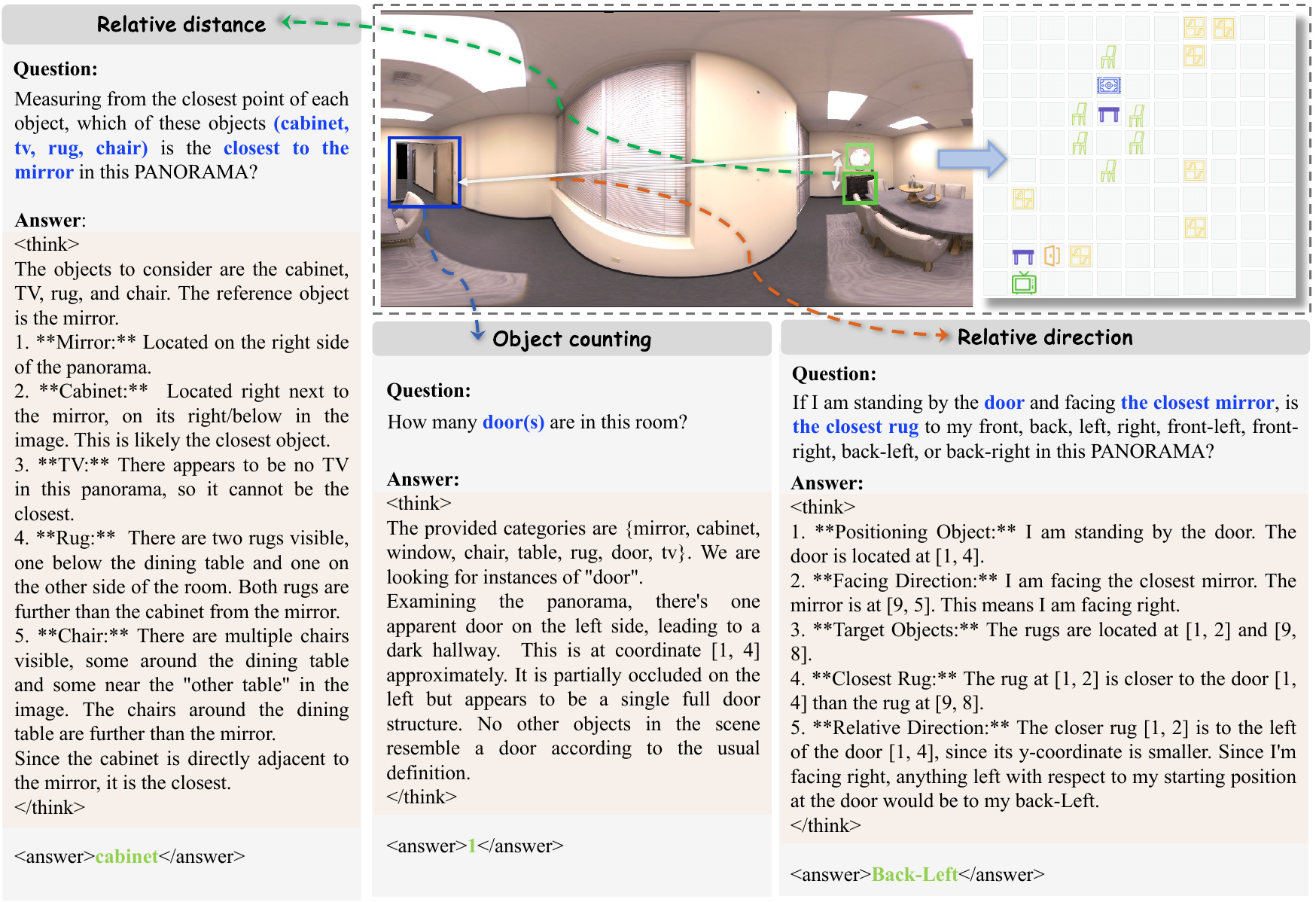}
\caption{Good case in think mode}
\label{think_mode}
\end{figure}

\end{document}